\documentclass[twoside]{article}

\usepackage[accepted]{aistats2026}
\usepackage[ruled,vlined,linesnumbered]{algorithm2e}

\usepackage[table]{xcolor}

\definecolor{bestcol}{RGB}{232,245,233} 
\newcommand{\bestcell}[1]{\cellcolor{bestcol}\textbf{#1}}
\newcommand{\bestdelta}[1]{\cellcolor{bestcol}\textbf{#1}}

\usepackage{array}
\usepackage{arydshln}
\usepackage{amsmath,amssymb,amsfonts,mathtools,bm}
\usepackage{amsthm}
\usepackage{microtype}
\usepackage{enumitem}
\usepackage{booktabs}
\usepackage{hyperref}

\usepackage[round]{natbib}

\begin{document}

%

%

\twocolumn[

\aistatstitle{CAWI: Copula-Aligned Weight Initialization for Randomized Neural Networks}

\aistatsauthor{Mushir Akhtar \And M. Tanveer \And  Mohd. Arshad }

\aistatsaddress{Indian Institute of Technology Indore, Simrol, Indore, 453552, India }
]

\begin{abstract}
  Randomized neural networks (RdNNs) enable efficient, backpropagation-free training by freezing randomly initialized input-to-hidden weights, which permits a closed-form solution for the output layer. However, conventional random initialization is blind to inter-feature dependence—ignoring correlations, asymmetries, and tail dependence in the data—which degrades conditioning and predictive performance. To the best of our knowledge, this limitation remains unaddressed in the RdNN literature. To close this gap, we propose CAWI (Copula-Aligned Weight Initialization), a framework that draws input-to-hidden weights from a data-fitted copula that matches empirical dependence, ensuring the frozen projections respect inter-feature dependence without sacrificing closed-form solution. CAWI (i) maps each feature to the unit interval using empirical CDFs, (ii) fits a multivariate copula that captures rank-based dependence among features, and (iii) samples each weight column $w_j$ from the fitted copula and applies a fixed inverse marginal transform to set scale. The objective, solver, and ``freeze-once'' paradigm remain unchanged; only the sampling law for $W$ becomes dependence-aware. For dependence modeling, we consider two copula families: elliptical (Gaussian, t) and Archimedean (Clayton, Frank, Gumbel). This enables CAWI to handle diverse dependence, including tail dependence. We evaluate CAWI across 83 diverse classification benchmarks (binary and multiclass) and two biomedical datasets, BreaKHis and the Schizophrenia dataset, using standard shallow and deep RdNN architectures. CAWI consistently delivers significant improvements in predictive performance over conventional random initialization. Codes are provided at \url{https://github.com/mtanveer1/CAWI}.
\end{abstract}

\section{Introduction}
Deep learning (DL) models have achieved remarkable success across many domains, from image recognition \cite{voulodimos2018deep, chen2019looks, pmlr-v235-liu24bq} to natural language processing \cite{9075398, lauriola2022introduction, pmlr-v235-yin24c}, due to their ability to learn rich hierarchical representations of data \cite{lecun2015deep, luo2023learning}. End-to-end training with backpropagation refines millions of parameters to minimize task-specific losses, yielding expressive features and strong generalization. However, state-of-the-art performance typically demands substantial computation, broad hyperparameter search, and long training cycles \cite{goodfellow2016deep, tiwari2023rcv2023}. Furthermore, training deep architectures can be hindered by vanishing or exploding gradient \cite{pmlr-v28-pascanu13, NEURIPS2022_31df5479, ceni2025random}.

The aforementioned limitations of DL models have prompted researchers to seek alternative neural architectures that couple competitive predictive performance with markedly lower training cost. A prominent family is randomized neural networks (RdNNs) \cite{pao1994learning, cao2018review, suganthan2021origins, 10220213}. In RdNNs, a substantial subset of parameters—typically input-to-hidden weights—are sampled once from simple distributions and then held fixed, while the remaining (often output) weights are obtained by solving a single linear system, commonly with Tikhonov regularization, instead of iterative backpropagation \cite{zhang2016survey}. This paradigm eliminates the need for iterative weight updates, substantially reducing training
time and computational overhead, yet still allows RdNNs to learn complex input-output mappings \cite{zhang2016survey}. Theoretically, RdNNs satisfy universal approximation on compact domains, approximating any continuous function arbitrarily well given sufficient width and appropriate activations \cite{igelnik1995stochastic, needell2024random}.

In recent years, RdNNs have witnessed significant advancements, particularly in methods aimed at improving their robustness and scalability. Ensemble-based approaches have been introduced to enhance generalization and mitigate variability \cite{shi2021random, 9447023}. Granular ball-based scalable methods have further extended the applicability of RdNNs to large-scale and high-dimensional datasets \cite{sajid2025gb}. Additionally, the integration of fuzzy inference systems into RdNN frameworks has provided a way to incorporate uncertainty handling and interpretability, making them more effective for complex and imprecise decision-making tasks \cite{10416391, 10552388}. Beyond these trends, a number of technical advances have broadened the RdNN toolkit \cite{wong2022online, chen2025adaptive, akhtar2025towards, ren2025foatlbl}.

Despite several advancements, RdNNs face a fundamental limitation rooted in the very mechanism that makes them efficient: the input-to-hidden weights are randomly initialized once and then held fixed throughout training. With the matrix \(W\in\mathbb{R}^{d\times h}\) frozen, the induced feature map \(x\mapsto \phi(xW+b)\) is non-adaptive; the hidden representation \(H=\phi(XW+\mathbf{1}_m b^\top)\in\mathbb{R}^{m\times h}\) is therefore determined by this random draw rather than learned from data.
The fixed representation \(H\) cannot adjust to dependencies among inputs (e.g., correlations, relative scales, higher-order interactions), so at practical hidden dimension \(h\) it may underrepresent the joint patterns that drive prediction. The readout then computes \(\hat{Y}=H\Theta\) with \(\Theta\in\mathbb{R}^{h\times n}\); if relevant dependencies are absent from \(H\), they cannot be recovered, leading to persistent approximation bias and lower performance at a given model size.

The preceding limitation naturally raises a question: can we make the frozen projections dependence-aware without abandoning closed-form training? In this work, we answer this question by introducing CAWI (Copula-Aligned Weight Initialization), a method that uses copula theory to generate \emph{dependence-aware} input-to-hidden weights. Copulas model multivariate data by separating the univariate marginals from the joint dependence structure, allowing us to learn dependence without committing to specific marginal forms \cite{nelsen2006introduction, joe2014dependence}. This separation is well suited to tabular learning, where feature scales and marginals vary widely but predictive signal often resides in cross-feature structure. Recent literature demonstrates broad utility of copulas in AI: copulas for time-series forecasting \cite{sun2022copula, ashok2024tactis}, interpretable estimation of CNN deep-feature densities using copulas and the generalized characteristic function \cite{chapman2024interpretable}, and survival analysis under dependent censoring with identifiability guarantees \cite{zhang2024deep}. Further developments include copula-nested spectral kernel networks \cite{tian2024copula}, scalable mixed models with arbitrary marginals \cite{simchoni2025flexible}, and neural copula density estimation \cite{letizia2025copula}, among others.

Several works in deep neural networks have explored structured or data-aware initialization strategies, such as orthogonal weight initialization~\cite{hu2020provable, narkhede2022review} and whitening-based transformations~\cite{lecun2002efficient, kessy2018optimal}. Orthogonal schemes stabilize signal propagation via algebraic constraints but remain distribution-agnostic, while whitening decorrelates features through covariance normalization, primarily addressing linear dependencies. More generally, data-dependent projection methods modify the feature space (e.g., via PCA or kernel mappings), often at additional computational cost.

In contrast, CAWI is designed for RdNNs, where hidden weights are sampled once and kept fixed. Rather than altering the feature representation or imposing structural constraints, CAWI modifies only the sampling distribution of the weight matrix using an empirical copula fitted to the training features. Specifically, we construct rank-based pseudo-observations, estimate elliptical (Gaussian, \(t\)) or Archimedean (Clayton, Frank, Gumbel) copulas, and draw each column of \(W\) from the fitted dependence structure with fixed marginal scaling. The resulting frozen projections inherit empirical inter-feature dependence, while the RdNN architecture, activation functions, objective, and closed-form solver remain unchanged, incurring essentially no additional training cost.

\textbf{Our Contributions}
\begin{itemize}
    \item We introduce CAWI (Copula-Aligned Weight Initialization), a plug-in scheme that makes randomized neural networks dependence-aware by sampling the columns of \(W\) from a copula fitted to the training features, while preserving the closed-form solver and the freeze-once paradigm.
    \item We provide numerically stable constructions for elliptical (Gaussian, \(t\)) and Archimedean (Clayton, Frank, Gumbel) copulas via rank-based estimation and nearest-SPD projection.
    \item We evaluate CAWI on 83 diverse classification benchmarks (binary and multiclass) and two biomedical datasets (BreaKHis and the Schizophrenia dataset) using both shallow and deep RdNN architectures, and observe consistent improvements, with 4--5\% relative gains on some datasets. 
\end{itemize}


\section{Preliminaries}

\subsection{Randomized neural network (RdNN) setting}
Let $X\in\mathbb{R}^{m\times d}$ be the input matrix ($m$ samples, $d$ features) and $Y\in\mathbb{R}^{m\times n}$ the target matrix (e.g., one–hot labels). An RdNN draws a single hidden layer of $h$ units by sampling a \emph{frozen} weight matrix $W \sim \mathcal{D}_W \;\subset\; \mathbb{R}^{d\times h},  b\in\mathbb{R}^{h}$, from a prescribed base distribution (classically i.i.d.\ uniform or Gaussian, independent across coordinates), and then \emph{never updates} these parameters. With an elementwise nonlinearity $\phi:\mathbb{R}\!\to\!\mathbb{R}$, the hidden representation is
\begin{equation}
H \;=\; \phi\!\big(XW + \mathbf{1}_m b^\top\big) \;\in\; \mathbb{R}^{m\times h},
\label{eq:rdnn-H}
\end{equation}
where $\mathbf{1}_m$ denotes an $m$–vector of ones broadcasting the bias. Training reduces to fitting only the output weights $\Theta\in\mathbb{R}^{h\times n}$ by ridge regression:
\begin{equation}
\Theta^\star
\;\in\;
\arg\min_{\Theta\in\mathbb{R}^{h\times n}}
\,\|H\Theta - Y\|_F^2 + \frac{\lambda}{2} \|\Theta\|_F^2,
\qquad \lambda\ge 0,
\label{eq:rdnn-obj}
\end{equation}
which admits the standard closed forms
\begin{align}
\text{(primal)}\quad
\Theta^\star &= (H^\top H + \lambda I_h)^{-1} H^\top Y,
\label{eq:rdnn-primal}\\
\text{(dual)}\quad
\Theta^\star &= H^\top (HH^\top + \lambda I_m)^{-1} Y.
\label{eq:rdnn-dual}
\end{align}
Thus \emph{no backpropagation is required}: once $W$ and $b$ are sampled, training is a single linear solve. Detailed formulation and architecture of standard RdNN models are discussed in Section~\ref{Section-A} of the Appendix.

\subsection{Copulas}
Copulas are a classical device for isolating and modeling \emph{dependence} separately from \emph{marginals} \cite{nelsen2006introduction, joe2014dependence}. Sklar’s seminal result shows that any multivariate distribution can be written as a coupling of its univariate marginals through a single function on the unit hypercube, thereby reducing multivariate modeling to estimating marginals and a dependence map on $[0,1]^d$ \cite{sklar1959fonctions}.

\begin{figure*}[t]
  \centering
  \includegraphics[width=\textwidth]{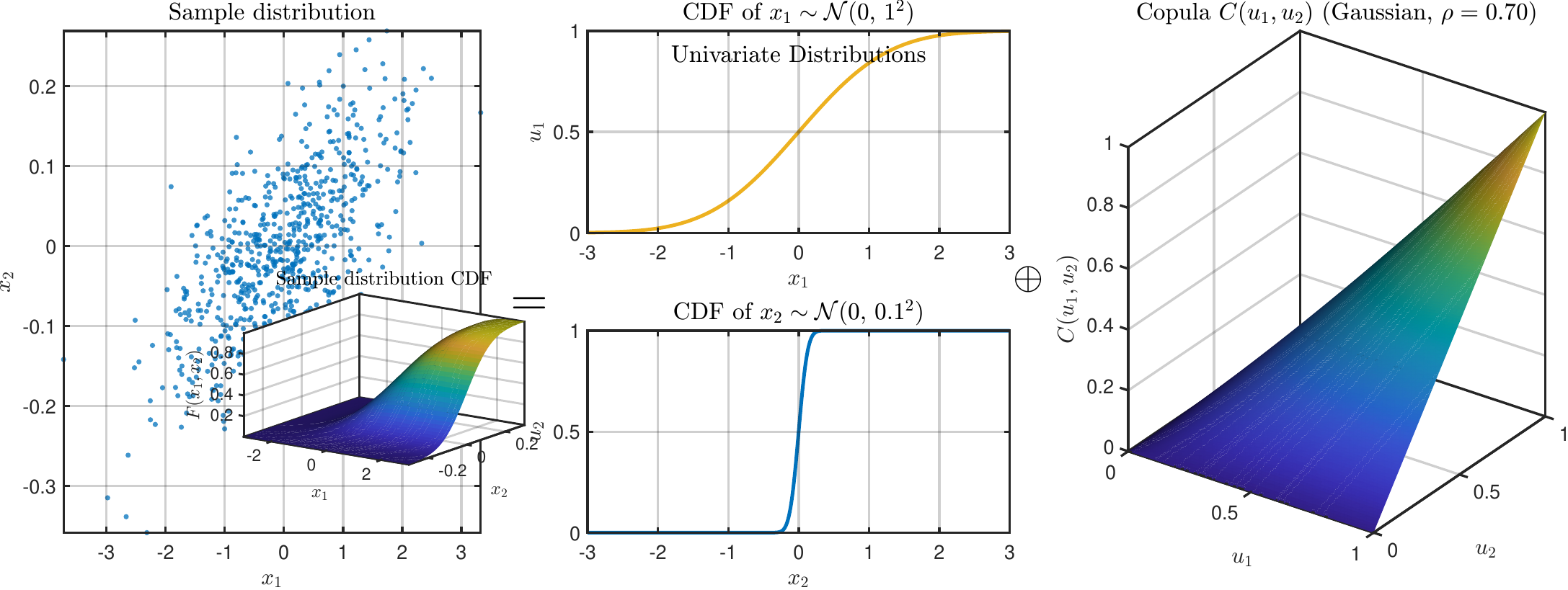}
  \caption{Schematic of a copula. \emph{Left:} empirical sample (with its joint CDF inset). 
  \emph{Middle:} univariate CDFs $F_1$ and $F_2$ (mapping $x_j \mapsto u_j=F_j(x_j)$). 
  \emph{Right:} the copula surface $C(u_1,u_2)$, i.e., the joint CDF on $[0,1]^2$ with uniform marginals. 
  The equality sign indicates Sklar’s factorization $F(x_1,x_2)=C(F_1(x_1),F_2(x_2))$; the $\oplus$ highlights that $C$ combines the marginal information into a dependence structure.}
  \label{fig:copula-schematic}
\end{figure*}

\paragraph{Copula:}
Let $\mathbf{X}=(X_1,\ldots,X_d)$ be a random vector with marginal cumulative distribution functions (CDFs) $F_j(t)=\Pr[X_j\le t]$ for $j=1,\ldots,d$. The \emph{copula associated with $\mathbf{X}$} is the joint CDF of the transformed vector $(F_1(X_1),\ldots,F_d(X_d))\in[0,1]^d$, namely
\begin{equation}
\begin{aligned}
C(u_1,\ldots,u_d)
&= \Pr\!\bigl\{ F_1(X_1)\le u_1,\ldots,F_d(X_d)\le u_d \bigr\},\\
&\quad (u_1,\ldots,u_d)\in[0,1]^d .
\end{aligned}
\end{equation}
Equivalently, a $d$–variate copula is a distribution function on $[0,1]^d$ whose univariate marginals are uniform on $(0,1)$. Figure~\ref{fig:copula-schematic} provides an intuitive picture of the concept.

\paragraph{Invariance:}
If $\psi_j$ are strictly increasing for each $j$, then $(\psi_1(X_1),\ldots,\psi_d(X_d))$ has the same copula as $(X_1,\ldots,X_d)$. Thus a copula depends only on cross-coordinate order, not on units or scales.

\paragraph{Sklar’s theorem:}
Let $F:\mathbb{R}^d\!\to[0,1]$ be the joint CDF of $\mathbf{X}=(X_1,\ldots,X_d)$ with marginals $F_1,\ldots,F_d$. There exists a copula $C$ such that, for all $\mathbf{x}=(x_1,\ldots,x_d)\in\mathbb{R}^d$,
\[
F(x_1,\ldots,x_d)
\;=\;
C\!\big(F_1(x_1),\ldots,F_d(x_d)\big).
\]
If each $F_j$ is continuous, the copula $C$ is unique. Conversely, for any copula $C$ and any marginals $F_1,\ldots,F_d$, the mapping above defines a valid $d$-variate distribution. If $X_1,\ldots,X_d$ are mutually independent, the copula is the \emph{product copula}
$C(u_1,\ldots,u_d)=\prod_{j=1}^d u_j$, representing absence of dependence among coordinates.

\section{Problem Statement}
Randomized neural networks (RdNNs) \emph{sample once and freeze} the input-to-hidden weight matrix \(W\in\mathbb{R}^{d\times h}\) and bias \(b\in\mathbb{R}^{h}\). With \(H=\phi(XW+\mathbf{1}_m b^\top)\) and \(\Theta^\star(W)=\arg\min_{\Theta}\{\|H\Theta-Y\|_F^2+\lambda\|\Theta\|_F^2\}\), training reduces to a single ridge-regularized linear solve---no backpropagation is required.

\paragraph{Limitation.}
The long\-standing bottleneck is how \(W\) is drawn: almost universally from an \emph{i.i.d.}, factorized base law (uniform or Gaussian). Equivalently, this imposes the \emph{product copula} \(C_{\perp}(u)=\prod_{j=1}^d u_j\) across weight coordinates, irrespective of the empirical dependence in the inputs \(X\) (correlations, asymmetries, tail dependence). The resulting mismatch can (i) reduce the diversity/effective rank of \(H\), (ii) fail to excite discriminative co-movements, and (iii) worsen the conditioning of \(H^\top H+\lambda I\).

\medskip\noindent
\textbf{Goal.} Replace the independence assumption on frozen weights \emph{without} altering the RdNN pipeline. Given data \(X\in\mathbb{R}^{m\times d}\), width \(h\), activation \(\phi\), and ridge \(\lambda\), training remains the same closed form; only the \emph{sampling law} for \(W\) changes. We seek a \emph{one-shot, data-aware} sampler \(\mathcal{S}\) that maps \(X\) to a distribution over \(W\) such that each column \(w\in\mathbb{R}^d\) has a joint law whose \emph{copula} matches an estimator of input dependence, while its marginals are simple and well-behaved:
\begin{equation}
\begin{aligned}
\operatorname{Copula}(w) &\approx \widehat{\operatorname{Copula}}(X), w_j \sim G~ \text{for a fixed marginal}~G\\
&\ (\text{e.g., }\mathcal{N}(0,1)\ \text{or}\ \mathcal{U}[-1,1]),\ j=1,\ldots,d.
\end{aligned}
\end{equation}

Biases \(b\) may be drawn independently from a simple one-dimensional law.

\medskip\noindent
\textbf{Design objective.} Among samplers satisfying the dependence-matching and marginal constraints, choose \(\mathcal{S}\) to improve downstream risk \emph{without} any iterative tuning of \(W\):
\begin{equation}
\begin{aligned}
\min_{\mathcal{S}}\;&
\mathbb{E}_{W\sim \mathcal{S}(X)}\,\mathbb{E}_{(x,y)\sim\mathcal{D}}
\big[\,\ell\big(h(x;W,b)\,\Theta^\star(W),\,y\big)\big] \\
\text{s.t.}\;&
\operatorname{Copula}(w)\approx \widehat{\operatorname{Copula}}(X),\quad
w_j \sim G,\ \ j=1,\ldots,d,\ 
\end{aligned}
\end{equation}
where \(h(x;W,b)=\phi(xW+b)\in\mathbb{R}^{h}\) and \(\mathcal{D}\) is the data distribution.
In short, the problem is to replace i.i.d.\ frozen weights with \emph{dependence-aware} frozen weights that align with the empirical copula of \(X\), thereby addressing the core weakness of random initialization while preserving the hallmark efficiency of RdNNs.

\section{Method}
We introduce \emph{Copula-Aligned Weight Initialization} (CAWI), a data-aware procedure that samples the frozen hidden weights in randomized neural networks using only the inputs \(X\). CAWI preserves the classical RdNN pipeline (no backpropagation; closed-form output layer) while aligning the \emph{dependence} among weight coordinates with that of the inputs. Concretely, CAWI (i) maps each feature to \([0,1]\) via its empirical cumulative distribution function to remove marginal scales, (ii) fits a multivariate copula on \([0,1]^d\) to summarize input dependence, and (iii) samples each weight column by drawing \(u\) from the fitted copula and transforming coordinates through a fixed marginal law \(G\) (e.g., \(\mathcal{N}(0,1)\) or \(\mathcal{U}[-1,1]\)) via inverse CDFs. The training objective and closed-form solver remain unchanged; only the law used to sample (and then freeze) \(W\) is replaced. The detailed CAWI procedure and its steps are presented next. Algorithm~\ref{alg:cawi} summarizes the procedure.

\paragraph{\textbf{Step 1: Probability-Integral Transform of Features}}
Given $X\in\mathbb{R}^{m\times d}$ with columns $X_{:j}$, we construct pseudo-observations
\[
U_{ij} \;=\; \widehat F_j(X_{ij}) \in (0,1), \qquad U\in (0,1)^{m\times d},
\]
where $\widehat F_j$ is the empirical cumulative distribution function (ECDF) of feature $j$, computed across samples (i.e., with $j$ fixed and $i=1,\dots,m$). A common rank-based implementation sets
$U_{ij} \;=\; \frac{\mathrm{rank}(X_{ij})}{m+1},$
where ranks are assigned feature-wise across samples. This guarantees $U_{ij}\in(0,1)$ (never exactly $0$ or $1$) and thus avoids infinities when applying inverse CDFs in Step 3. 

This transformation removes units and strictly monotone re-scalings, isolating cross-feature \emph{dependence} in the joint distribution of $U$, consistent with standard empirical copula construction.

\paragraph{\textbf{Step 2: Fit a Multivariate Copula on $U$}}
Let $\mathcal{C}=\{\,C(\cdot;\eta):\eta\in\Xi\,\}$ be a parametric family of $d$-variate copulas on $[0,1]^d$, with parameter $\eta$ in parameter space $\Xi$. Using the pseudo-observations $U$, we estimate $\widehat\eta$ to obtain a fitted copula
$\widehat C(\cdot)\;=\;C(\cdot;\widehat\eta),$
which summarizes the empirical dependence of the inputs and has uniform marginals by construction. Estimation is rank-based: one may maximize the copula pseudo-likelihood, solve method-of-moments equations (e.g., matching Kendall’s $\tau$), or employ composite likelihoods over pairs when $d$ is large. This fit is performed \emph{once} per training split and is agnostic to the downstream task. The procedure is copula-agnostic: any $d$-variate copula family for which an estimator and a sampler are available (elliptical or Archimedean) can be plugged into $\mathcal{C}$. Concrete instantiations used in our work are detailed in \ref{subsec:copula-families}.

\paragraph{\textbf{Step 3: Sample Dependence-Aligned Weight Columns}}
Choose a continuous one-dimensional marginal law $G$ with a strictly increasing quantile $G^{-1}$ for individual weight entries (e.g., $\mathcal N(0,1)$ or $\mathcal U[-1,1]$). 
For each hidden unit $t=1,\dots,h$:
\begin{enumerate}
\item draw $u^{(t)} \sim \widehat C$ in $[0,1]^d$;
\item set $w^{(t)} \gets G^{-1}\!\big(u^{(t)}\big)$ coordinatewise, i.e., $w^{(t)}_j \gets G^{-1}(u^{(t)}_j)$ for $j=1,\ldots,d$;
\item assign the $t^{th}$ column of $W$ as $W_{:t} \gets w^{(t)}$ and draw a scalar bias $b_t$ independently from a simple one-dimensional law.
\end{enumerate}
Since $G^{-1}$ is strictly increasing for continuous $G$, the \emph{joint law of the coordinates of} $w^{(t)}$ has copula $\widehat C$ while its univariate marginals are $G$, thereby matching input dependence while retaining convenient weight marginals.

\paragraph{\textbf{Step 4: Closed-Form Training (Unchanged)}}
With the dependence-aligned weights \(W\), we form the hidden matrix $H \;=\; \phi\!\big(XW + \mathbf{1}_m b^\top\big)$,
and obtain the output layer by the standard ridge-regularized least squares
\[
\Theta^\star \;=\; \arg\min_{\Theta}\ \|H\Theta - Y\|_F^2 + \frac{\lambda}{2} \|\Theta\|_F^2.
\]
No iterative updates of \(W\) are introduced: CAWI modifies only the sampling law used to freeze the hidden weights; the training and inference pipeline remains identical to a classical RdNN.

\begin{algorithm}[t]
\caption{CAWI: Copula–Aligned Weight Initialization for Randomized Neural Networks}
\label{alg:cawi}
\DontPrintSemicolon
\SetKwInOut{Input}{Input}\SetKwInOut{Output}{Output}
\Input{Data $X\in\mathbb{R}^{m\times d}$, width $h$, activation $\phi$, ridge $\lambda$, weight marginal $G$, copula family $\mathcal{C}=\{C(\cdot;\eta):\eta\in\Xi\}$.}
\Output{Frozen weights $W\in\mathbb{R}^{d\times h}$, bias vector $b\in\mathbb{R}^{h}$, output weights $\Theta^\star$.}

\BlankLine
\tcp{Step 1: Probability$-$integral transform (pseudo$-$observations)}
\For{$j \gets 1$ \KwTo $d$}{
  compute ECDF $\widehat F_j$ of column $X_{:j}$\;
  \For{$i \gets 1$ \KwTo $m$}{ $U_{ij} \gets \widehat F_j(X_{ij}) \in (0,1)$ }
}
\BlankLine

\tcp{Step 2: Fit a $d-$variate copula on $U$ (details in \ref{subsec:copula-families})}
Estimate $\widehat\eta$ from $U$ (rank–based); set $\widehat C(\cdot) \gets C(\cdot;\widehat\eta)$\;
\BlankLine

\tcp{Step 3: Sample dependence$-$aligned weight columns}
\For{$t \gets 1$ \KwTo $h$}{
  draw $u^{(t)} \sim \widehat C$ in $[0,1]^d$\;
  \For{$j \gets 1$ \KwTo $d$}{ $w^{(t)}_j \gets G^{-1}\!\big(u^{(t)}_j\big)$ }
  set $W_{:t} \gets w^{(t)}$\;
}
Sample each bias $b_t$ independently from a simple one–dimensional law (e.g., uniform), for $t=1,\dots,h$\;
\BlankLine

\tcp{Step 4: Closed$-$form output layer (unchanged)}
Form hidden features $H \gets \phi\!\big(XW + \mathbf{1}_m b^\top\big)$\;
Compute $\Theta^\star$\;

\end{algorithm}

\subsection{Copula Families: Parameterization and Estimation}
\label{subsec:copula-families}
We instantiate Step~2 with five widely used \(d\)-variate copula families that together capture symmetric vs.\ asymmetric dependence and both light- and heavy-tailed co-movement: \emph{Gaussian} (elliptical, tail-neutral), \emph{\(t\)} (elliptical, heavy-tailed), \emph{Clayton} (Archimedean, lower-tail dependent), \emph{Gumbel} (Archimedean, upper-tail dependent), and \emph{Frank} (Archimedean, tail-neutral, non-elliptical). This selection balances expressive coverage with tractable rank-based estimation and efficient sampling, enabling CAWI to adapt to diverse empirical dependence while keeping the training pipeline simple and stable.

\paragraph{\textbf{Elliptical families (Gaussian, $t$)}}
Both families are parameterized by a correlation matrix \(R\in\mathbb{R}^{d\times d}\); the \(t\) copula additionally has degrees of freedom \(\nu>2\). We estimate \(R\) from pairwise Kendall’s \(\tau\) computed on \(U\) via the elliptical identity
\[
\tau_{ij} \;=\; \tfrac{2}{\pi}\arcsin(R_{ij})
\Rightarrow
\widehat R_{ij} \;=\; \sin\!\big(\tfrac{\pi}{2}\,\widehat\tau_{ij}\big)
\text{(elementwise)},
\]
followed by a nearest-correlation projection to enforce symmetry, unit diagonal, and positive (semi)definiteness (a vanishing ridge may be added if needed for numerical stability). For the \(t\) copula, we then profile a rank-based pseudo-likelihood over \(\nu\) with \(R\) fixed to obtain \(\widehat\nu\) (a one-dimensional search). The fitted copula \(\widehat C\) yields draws \(U\in(0,1)^{N\times d}\), which are mapped to weights through the inverse marginals in Step~3.

\paragraph{\textbf{Archimedean families (Clayton, Gumbel, Frank)}}
These families are governed by a single scalar parameter \(\theta\) that controls concordance and tail behavior. We employ robust, composite, rank-based estimation from pairwise Kendall’s \(\tau\) computed on \(U\). Let \(\widehat\tau_{ij}\) denote the sample Kendall’s \(\tau\) between coordinates \(i\) and \(j\), and let \(\bar\tau\) be their average over \(i<j\). With this setup, we estimate a single $\theta$ per family as follows.

\begin{itemize}
\item \textbf{Clayton} (\(\theta>0\), lower-tail dependent). The \(\tau\)–\(\theta\) relation \(\tau=\theta/(\theta+2)\) yields
\[
\widehat\theta \;=\; \frac{2\,\bar\tau}{1-\bar\tau}\,,
\]
or, equivalently, one may minimize a composite method-of-moments objective \(\sum_{i<j}\!\big(\tfrac{\theta}{\theta+2}-\widehat\tau_{ij}\big)^2\).
\item \textbf{Gumbel} (\(\theta\ge 1\), upper-tail dependent). With \(\tau=1-\tfrac{1}{\theta}\),
\[
\widehat\theta \;=\; \frac{1}{1-\bar\tau}\,,
\]
or via the same composite fit \(\sum_{i<j}\!\big(1-\tfrac{1}{\theta}-\widehat\tau_{ij}\big)^2\).
\item \textbf{Frank} (\(\theta\in\mathbb{R}\setminus\{0\}\), tail-neutral, non-elliptical). Kendall’s \(\tau\) satisfies
\[
\tau(\theta)\;=\;1-\frac{4}{\theta}+\frac{4}{\theta}\,D_1(\theta),
\]
where \(D_1\) is the Debye function of order~1. We obtain \(\widehat\theta\) by numerically inverting \(\tau(\theta)\) at \(\bar\tau\) (monotone one-dimensional root-finding), or by composite pseudo-likelihood.
\end{itemize}
For \(d>2\) we adopt the standard exchangeable extension, i.e., a single global parameter \(\widehat\theta\) shared across dimensions. This delivers stable, scalable \(d\)-variate fits and admits efficient generator-based sampling of \(U\in(0,1)^{N\times d}\). As in Step~3, the sampled \(U\) is then mapped coordinatewise through the chosen marginal quantile \(G^{-1}\) to produce dependence-aligned weight columns.

\subsection{Computational Complexity of CAWI}\label{subsec:computational_complexity}

CAWI adds a one-time preprocessing stage (Steps 1-3) on top of the RdNN pipeline. Let \(m\) denote the number of samples, \(d\) the input dimensionality, and \(h\) the width (number of hidden units).

\paragraph{Step 1: Probability-integral transform.}
Computing the empirical CDF or rank transform for each of the \(d\) feature columns requires sorting \(m\) values per column, leading to a total cost of \(O(d\, m \log m)\).

\paragraph{Step 2: Copula fitting.}
Using rank-based Kendall's \(\tau\) for dependence estimation requires computing pairwise concordance statistics for all \(d(d-1)/2 = O(d^2)\) feature pairs. A standard implementation of Kendall's \(\tau\) costs \(O(m^2)\) per pair, yielding an overall complexity of \(O(d^2 m^2)\).  
For elliptical copulas (Gaussian or \(t\)), an additional nearest-correlation or spectral projection step contributes \(O(d^3)\). The \(t\)-copula further requires a one-dimensional search over degrees of freedom, with per-iteration cost \(O(m d^2)\), which remains negligible relative to the dominant \(O(d^2 m^2)\) term.  For Archimedean copulas (Clayton, Gumbel, Frank), estimating a single scalar parameter from averaged Kendall’s \(\bar{\tau}\) incurs negligible cost beyond the \(O(d^2 m^2)\) computation of the pairwise \(\tau\)-values.

\paragraph{Step 3: Sampling dependence-aligned weight columns.}
For elliptical copulas, sampling requires an initial factorization of the \(d \times d\) correlation matrix, costing \(O(d^3)\), followed by \(h\) matrix-vector multiplications, each costing \(O(d^2)\), for a total of \(O(d^3 + h d^2)\).  
For Archimedean copulas, sampling each column requires only \(O(d)\) operations, yielding a total cost of \(O(h d)\).  
In both cases, mapping the copula samples through the marginal quantile \(G^{-1}\) costs \(O(h d)\), which is dominated by the previous terms.

\paragraph{Step 4: Closed-form RdNN training (unchanged).}
With dependence-aligned frozen weights, hidden features are formed as \(H = \phi(XW + \mathbf{1}_m b^\top)\). The cost of forming \(XW\) is \(O(m d h)\), and computing the ridge-regularized output layer using the normal equations costs \(O(m h^2 + h^3)\). This is identical to classical RdNNs.

\paragraph{Summary.}
CAWI introduces a one-time preprocessing cost of  $O\big(d\ m \log m + d^2 m^2 + d^3 + h d^2\big)$, while the dominant runtime in practice remains the standard RdNN training step,  
$O(m d h + m h^2 + h^3)$.
Thus, CAWI preserves the hallmark efficiency of RdNNs.
The empirical training-time analysis reported in Section~E of the Appendix further confirms this efficiency.

\section{Empirical Results}
We evaluate CAWI by integrating it into representative randomized architectures spanning \emph{shallow}, \emph{deep}, and \emph{wide} settings:
\emph{(i) RVFL} (single hidden layer with direct links) \cite{pao1994learning},
\emph{(ii) dRVFL} (deep RVFL) \cite{shi2021random}, and
\emph{(iii) BLS} (Broad Learning System; a wide, block-structured random-feature architecture) \cite{chen2017broad}. We compare the standard i.i.d.\ baseline against CAWI variants that differ only in the fitted copula used to sample columns of $W$:
\begin{itemize}
\item \textbf{i.i.d.\ baseline}: \(W\) is sampled independently from a Uniform distribution (\(\mathcal{U}[-1,1]\)).
\item \textbf{CAWI (elliptical)}: \(W\) is sampled using \emph{Gaussian} and \emph{$t$} copulas.
\item \textbf{CAWI (Archimedean)}: \(W\) is sampled using \emph{Clayton}, \emph{Frank}, and \emph{Gumbel} copulas.
\end{itemize}
The detailed experimental setup is provided in Section~\ref{Section-B} of the Appendix.
\begin{table*}[]
\centering
\caption{Results on the BreaKHis dataset for RVFL baseline and copula-based initializations (test accuracy, \%). Best per row is \textbf{bold} and highlighted; the last column reports the improvement of the best method over the i.i.d.\ baseline.}
\label{tab:rvfl-breakhis-accuracy}
\begin{tabular}{l|c:ccccc:c}
\toprule
\textbf{Dataset} & \textbf{iid} & \textbf{Gaussian} & \(\boldsymbol{t}\) & \textbf{Clayton} & \textbf{Frank} & \textbf{Gumbel} & \textbf{Improvement} \\
\midrule
AD vs.\ DC & 71.0138 & \bestcell{73.2616} & 72.3041 & 69.7440 & 70.0666 & 71.0292 & \bestdelta{+2.2478} \\
AD vs.\ LC & 62.1429 & 62.9507 & \bestcell{65.0425} & 62.9932 & 63.7500 & 63.4099 & \bestdelta{+2.8996} \\
AD vs.\ MC & 63.6364 & 65.0909 & \bestcell{66.5455} & 64.3636 & 65.0909 & 65.4545 & \bestdelta{+2.9091} \\
AD vs.\ PC & 60.2211 & 61.4796 & 62.2959 & 61.4371 & 62.3044 & \bestcell{62.7126} & \bestdelta{+2.4915} \\
FA vs.\ DC & 63.5955 & \bestcell{65.3933} & \bestcell{65.3933} & \bestcell{65.3933} & 64.2697 & 64.2697 & \bestdelta{+1.7978} \\
FA vs.\ LC & 65.7766 & 66.3027 & \bestcell{67.3622} & 66.0324 & 66.0396 & 66.8432 & \bestdelta{+1.5856} \\
FA vs.\ MC & 59.8464 & 59.6025 & 60.3403 & 60.8461 & \bestcell{61.0720} & 60.5932 & \bestdelta{+1.2256} \\
FA vs.\ PC & 68.0000 & 67.7333 & \bestcell{69.0667} & 66.6667 & 67.7333 & 67.4667 & \bestdelta{+1.0667} \\
PT vs.\ DC & 67.1731 & 70.2933 & 71.1971 & \bestcell{71.8125} & 69.9760 & 70.2933 & \bestdelta{+4.6394} \\
PT vs.\ LC & 57.0980 & \bestcell{62.2980} & 60.2824 & 59.0510 & 59.0980 & 60.2667 & \bestdelta{+5.2000} \\
PT vs.\ MC & 61.9987 & 62.3120 & 61.9799 & 61.6228 & \bestcell{63.7343} & 60.9398 & \bestdelta{+1.7356} \\
PT vs.\ PC & 57.3569 & 58.1098 & 59.6941 & 58.5098 & \bestcell{59.7176} & 58.1098 & \bestdelta{+2.3607} \\
TA vs.\ DC & 65.3995 & 65.9877 & 66.5847 & \bestcell{67.1817} & 66.8745 & 66.8832 & \bestdelta{+1.7822} \\
TA vs.\ LC & 62.1663 & 62.8931 & \bestcell{63.6897} & 63.6338 & 62.5577 & 62.5646 & \bestdelta{+1.5234} \\
TA vs.\ MC & 57.1751 & 59.5311 & 59.8418 & 58.5254 & \bestcell{60.8701} & 58.5537 & \bestdelta{+3.6950} \\
TA vs.\ PC & 56.3382 & 56.7296 & 58.5674 & 57.4284 & 59.7065 & \bestcell{61.9427} & \bestdelta{+5.6045} \\
\bottomrule
\end{tabular}
\end{table*}

\begin{table*}[]
\centering
\caption{Results on the BreaKHis dataset for dRVFL baseline and copula-based initializations (test accuracy, \%). Best per row is \textbf{bold} and highlighted; the last column reports the improvement of the best method over the i.i.d.\ baseline.}
\label{tab:drvfl-breakhis-accuracy}
\begin{tabular}{l|c:ccccc:c}
\toprule
\textbf{Dataset} & \textbf{iid} & \textbf{Gaussian} & \(\boldsymbol{t}\) & \textbf{Clayton} & \textbf{Frank} & \textbf{Gumbel} & \textbf{Improvement} \\
\midrule
AD vs.\ DC & 74.1987 & 71.0087 & 69.7542 & 66.2366 & \bestcell{75.2514} & 66.2366 & \bestdelta{+1.0527} \\
AD vs.\ LC & 62.5595 & 61.7602 & 61.3350 & \bestcell{67.5170} & 65.8503 & 61.3180 & \bestdelta{+4.9575} \\
AD vs.\ MC & 67.2727 & 63.6364 & 65.0909 & 67.7273 & \bestcell{68.9091} & 67.0909 & \bestdelta{+1.6364} \\
AD vs.\ PC & 65.9949 & 56.5391 & 56.5391 & 56.9473 & \bestcell{68.9541} & 56.9473 & \bestdelta{+2.9592} \\
FA vs.\ DC & 66.5169 & 64.7191 & 66.0674 & \bestcell{67.6338} & 67.5577 & 67.5646 & \bestdelta{+1.1169} \\
FA vs.\ LC & 69.7802 & 67.9063 & 67.6505 & 68.1802 & 68.4505 & \bestcell{70.0541} & \bestdelta{+0.2739} \\
FA vs.\ MC & 58.3740 & 58.3740 & 59.1117 & 58.3740 & 58.3740 & \bestcell{60.5781} & \bestdelta{+2.2041} \\
FA vs.\ PC & 72.8000 & 70.2933 & 73.1971 & \bestcell{74.8125} & 72.9760 & 73.2933 & \bestdelta{+2.0125} \\
PT vs.\ DC & 64.4135 & 72.4375 & 68.7260 & 64.4135 & \bestcell{77.3846} & 64.4135 & \bestdelta{+12.9711} \\
PT vs.\ LC & 55.8980 & 57.4667 & 56.3137 & \bestcell{57.8824} & 57.0980 & 56.2745 & \bestdelta{+1.9844} \\
PT vs.\ MC & 61.9862 & 60.2130 & 60.5576 & \bestcell{65.8521} & 61.2657 & 60.5576 & \bestdelta{+3.8659} \\
PT vs.\ PC & 54.9725 & 54.5725 & \bestcell{57.7412} & 55.3804 & 54.9725 & 56.1725 & \bestdelta{+2.7687} \\
TA vs.\ DC & 61.5496 & 61.5496 & 61.5496 & \bestcell{62.2137} & 61.5496 & 61.5496 & \bestdelta{+0.6641} \\
TA vs.\ LC & 68.1831 & 63.6408 & 63.7107 & \bestcell{69.2802} & 67.0510 & 65.9329 & \bestdelta{+1.0971} \\
TA vs.\ MC & 57.1751 & 56.8418 & 56.8418 & \bestcell{60.8531} & 56.8418 & 56.8418 & \bestdelta{+3.6780} \\
TA vs.\ PC & 57.0860 & 58.6164 & \bestcell{61.1810} & 57.8407 & 57.4563 & 57.0720 & \bestdelta{+4.0950} \\
\bottomrule
\end{tabular}
\end{table*}
\begin{table*}[]
\centering
\caption{Results on the Schizophrenia ROI datasets for RVFL baseline and copula-based initializations (test accuracy, \%). Best per row is \textbf{bold} and highlighted; the last column reports the improvement of the best method over the i.i.d.\ baseline.}
\label{tab:rvfl-schizo-accuracy}
\begin{tabular}{l|c:ccccc:c}
\toprule
\textbf{Dataset} & \textbf{iid} & \textbf{Gaussian} & \(\boldsymbol{t}\) & \textbf{Clayton} & \textbf{Frank} & \textbf{Gumbel} & \textbf{Improvement} \\
\midrule
SZ--ROI GM+WM & 60.3448 & 61.7701 & 63.0805 & 63.1034 & 62.4138 & \bestcell{63.1264} & \bestdelta{+2.7816} \\
SZ--ROI GM    & 62.4138 & \bestcell{63.7931} & 62.4368 & 62.4598 & 61.7931 & 61.1264 & \bestdelta{+1.3793} \\
SZ--ROI WM    & 61.0805 & 60.3448 & 61.0575 & \bestcell{62.3908} & 61.0345 & 59.6782 & \bestdelta{+1.3103} \\
\bottomrule
\end{tabular}
\end{table*}

\begin{table*}[]
\centering
\caption{Results on the Schizophrenia ROI datasets for dRVFL baseline and copula-based initializations (test accuracy, \%). Best per row is \textbf{bold} and highlighted; the last column reports the improvement of the best method over the i.i.d.\ baseline.}
\label{tab:drvfl-schizo-accuracy}
\begin{tabular}{l|c:ccccc:c}
\toprule
\textbf{Dataset} & \textbf{iid} & \textbf{Gaussian} & \(\boldsymbol{t}\) & \textbf{Clayton} & \textbf{Frank} & \textbf{Gumbel} & \textbf{Improvement} \\
\midrule
SZ--ROI GM+WM & 71.2874 & \bestcell{74.7816} & 72.6437 & 71.9770 & 71.9310 & 72.6437 & \bestdelta{+3.4942} \\
SZ--ROI GM    & 68.5287 & 69.3103 & 70.5747 & 68.5287 & \bestcell{70.6897} & 68.5747 & \bestdelta{+2.1610} \\
SZ--ROI WM    & 63.7701 & 64.4828 & 67.1494 & \bestcell{67.2184} & 65.8391 & 66.5287 & \bestdelta{+3.4483} \\
\bottomrule
\end{tabular}
\end{table*}
\subsection{Dataset Description} The evaluation is conducted on 83 benchmark datasets downloaded from the UCI repository \cite{dua2017uci}, comprising 30 binary and 53 multiclass classification tasks. The number of classes ranges from 2 to 26. Sample sizes span from 24 to 58{,}000 instances, and the number of features ranges from 3 to 263, covering both small- and large-scale datasets as well as low- and high-dimensional regimes. Detailed dataset statistics are provided in Table~\ref{tab:dataset_detail} of the Appendix. To validate performance in real-world settings, we additionally evaluate CAWI on two biomedical datasets: the \emph{BreaKHis} breast cancer dataset \cite{spanhol2015dataset} and \emph{Schizophrenia ROI} dataset from the COBRE repository (\url{http://fcon_1000.projects.nitrc.org/indi/retro/cobre.html}). 
For BreaKHis, we use 1,240 histopathological image scans at \(400\times\) magnification, organized into benign and malignant classes. The benign subclasses are adenosis (AD, 106), fibroadenoma (FA, 237), phyllodes tumor (PT, 115), and tubular adenoma (TA, 130); the malignant subclasses are lobular carcinoma (LC, 137), papillary carcinoma (PC, 138), ductal carcinoma (DC, 208), and mucinous carcinoma (MC, 169). Feature extraction follows the procedure in \cite{gautam2020minimum}. 
The Schizophrenia ROI dataset includes 72 schizophrenia subjects (ages 18--65; mean \(38.1\pm13.9\) years) and 74 healthy controls (ages 18--65; mean \(35.8\pm11.5\) years); ROI features are derived following \cite{tanveer2022intuitionistic}.

\subsection{Performance Evaluation}

In this subsection, we present empirical results on three groups of benchmarks: the BreaKHis dataset, the Schizophrenia ROI dataset, and a broader suite of 83 UCI benchmark datasets. For BreaKHis and Schizophrenia ROI, we evaluate RVFL, dRVFL, and BLS under the standard i.i.d.\ initialization and the proposed CAWI variants (Gaussian, $t$, Clayton, Frank, and Gumbel). Tables~\ref{tab:rvfl-breakhis-accuracy} and~\ref{tab:drvfl-breakhis-accuracy} report the BreaKHis results for RVFL and dRVFL, while Tables~\ref{tab:rvfl-schizo-accuracy} and~\ref{tab:drvfl-schizo-accuracy} present the corresponding results for the Schizophrenia ROI dataset. Results for BLS are provided in Tables~\ref{tab:bls-breakhis-accuracy}--\ref{tab:bls-schizo-accuracy} of the Appendix. For the broader benchmark evaluation, we report results on 83 UCI datasets (30 binary and 53 multiclass). The aggregate summary is presented in Table~\ref{tab:uci-aggregate-summary}, while the complete per-dataset results are available in Tables~\ref{tab:rvfl-benchmark-accuracy}--\ref{tab:rvfl-multiclass-accuracy} of the Appendix. All these results are reported using a fixed seed (\texttt{rng(42, `twister')}); additional experiments with two independent seeds (7 and 123) are provided in Section~\ref{Section-C} of the Appendix to assess seed sensitivity.

\subsubsection{Result on BreaKHis Dataset}Table~\ref{tab:rvfl-breakhis-accuracy} shows that \emph{every} pairwise task attains a higher test accuracy with a copula-aligned initialization than with the i.i.d.\ baseline (rightmost column $>0$ for all 16 rows). Gains range from about $+1.07$ to $+5.60$ percentage points (e.g., TA vs.\ PC: $+5.60$, PT vs.\ LC: $+5.20$, PT vs.\ DC: $+4.64$), indicating that dependence-aware sampling of \(W\) provides a meaningful lift even in a shallow RVFL. Not every copula family dominates on every task; rather, the \emph{best} variant shifts across pairs, consistent with heterogeneous dependence patterns in the features. Across the 16 pairs, the winning family varies: the $t$-copula is best on 5 tasks (AD vs.\ LC/MC; FA vs.\ LC; TA vs.\ LC; plus a three-way tie in FA vs.\ DC), \emph{Gaussian} on 3 (e.g., AD vs.\ DC; FA vs.\ PC; PT vs.\ LC), \emph{Clayton} on 2 (PT vs.\ DC; TA vs.\ DC), \emph{Frank} on 4 (FA vs.\ MC; PT vs.\ MC/PC), and \emph{Gumbel} on 2 (AD vs.\ PC; TA vs.\ PC). A three-way tie (Gaussian/$t$/Clayton) in FA vs.\ DC suggests that distinct copula families can approximate the relevant dependence equally well in some cases. Overall, every BreaKHis pair benefits from copula-aligned initialization (all improvements $>0$), with gains up to $+5.60$ percentage points, reinforcing the central claim: injecting empirical dependence into the frozen projections consistently improves RVFL performance, while different copula families provide complementary advantages across tasks.
Table~\ref{tab:drvfl-breakhis-accuracy} shows that copula-aligned initialization again improves over the i.i.d.\ baseline on \emph{every} pair (rightmost column $>0$ for all 16 rows), now with a wider spread of gains: from $+0.27$ to $+12.97$ percentage (e.g., PT vs.\ DC: $+12.97$, TA vs.\ PC: $+4.10$, PT vs.\ MC: $+3.87$). This indicates that dependence-aware sampling remains beneficial in deeper randomized stacks and can yield larger lifts when multiple frozen layers compound representational biases. The winning family varies by pair, with a clear pattern favoring Archimedean copulas: \emph{Clayton} leads on \textbf{8} tasks, \emph{Frank} on \textbf{4}, \emph{Gumbel} on \textbf{2}, and the \emph{$t$}-copula on \textbf{2}; \emph{Gaussian} does not take a win in this setting. These outcomes are consistent with heterogeneous, often asymmetric dependence in the extracted features: Clayton/Frank/Gumbel (which capture different tail/asymmetry profiles) dominate, while elliptical models win when dependence is more symmetric.

\subsubsection{Result on Schizophrenia Dataset}
Across the three ROI configurations (GM+WM, GM, WM), CAWI improves upon the i.i.d.\ baseline in both architectures. 
In the RVFL setting (see Table ~\ref{tab:rvfl-schizo-accuracy}), the best copula varies by ROI: \emph{Gumbel} leads on GM+WM (\(+2.78\)), \emph{Gaussian} on GM (\(+1.38\)), and \emph{Clayton} on WM (\(+1.31\)). 
For dRVFL (see Table ~\ref{tab:drvfl-schizo-accuracy}), the pattern shifts: \emph{Gaussian} tops GM+WM (\(+3.49\) pp), \emph{Frank} on GM (\(+2.16\)), and \emph{Clayton} again wins on WM (\(+3.45\)). 
Overall, these results echo the BreaKHis findings without redundancy: dependence-aware sampling of \(W\) consistently lifts accuracy, and the most effective copula family depends on the ROI subset and architecture—underscoring the value of modeling the \emph{type} of dependence (symmetric vs.\ asymmetric; central vs.\ tail) rather than assuming independence.

\subsection{Results on 83 UCI Datasets}

On the broader benchmark suite, we evaluate RVFL on 83 UCI datasets. To provide a concise and transparent summary of the empirical findings, we report the aggregate performance of CAWI in Table~\ref{tab:uci-aggregate-summary}. For each dataset, we compare the conventional i.i.d.\ initialization with the best-performing CAWI configuration (i.e., the copula family yielding the highest test accuracy for that dataset). This reflects the intended usage of CAWI as a flexible, copula-agnostic framework capable of adapting to dataset-specific dependency structures. Across all 83 datasets, the best CAWI variant outperforms the i.i.d.\ baseline on every dataset. The average test accuracy of the i.i.d.\ initialization is $78.08\%$, whereas the average accuracy of the best CAWI configuration reaches $79.35\%$, corresponding to a mean improvement of $+1.27$ percentage points. These results demonstrate that CAWI yields consistent improvements across both binary and multiclass problems, rather than isolated gains on a small subset of datasets. 

\begin{table}[t]
\centering
\caption{Aggregate performance summary across 83 UCI datasets.}
\label{tab:uci-aggregate-summary}
\begin{tabular}{lc}
\toprule
\textbf{Metric} & \textbf{Value} \\
\midrule
Mean accuracy (i.i.d.) & 78.08\% \\
Mean accuracy (Best CAWI) & 79.35\% \\
Mean improvement & +1.27\% \\
Datasets where CAWI $>$ i.i.d. & 83 / 83 \\
\bottomrule
\end{tabular}
\end{table}

\section{Conclusions}
In this work, we address a foundational gap in randomized neural networks, namely that input-to-hidden weights are randomly initialized and then held fixed throughout training. We present CAWI, a plug-in initialization scheme that aligns the frozen weight columns with the empirical dependence structure of the inputs via a fitted copula, while leaving the classical RdNN pipeline—closed-form readout, no backpropagation—unchanged. By modifying only the sampling law for \(W\), CAWI yields dependence-aware hidden projections with a computational footprint comparable to i.i.d.\ initialization. Building on this design, our empirical study shows consistent and often substantial gains across 83 UCI benchmarks and two biomedical datasets (BreaKHis and Schizophrenia ROI) using RVFL, dRVFL, and BLS architectures, all while preserving backpropagation-free training. Codes are provided at \url{https://github.com/mtanveer1/CAWI}.

A natural next step for future research is the development of automatic copula selection strategies. Rather than manually evaluating multiple copula families, a data-driven mechanism could identify the most suitable dependency structure for a given dataset and initialize the model accordingly. Such an adaptive selection procedure would further streamline CAWI while preserving its computational efficiency and enhancing its ability to capture dataset-specific dependence patterns.

\section*{Acknowledgement}

Mushir Akhtar acknowledges the financial support received from the CSIR, New Delhi, under Fellowship Grant No.~09/1022(13849)/2022-EMR-I. Mohd. Arshad acknowledges the funding support provided by the ANRF, India, through the Core Research Grant (Grant No.~CRG/2023/001230).

\bibliography{refs}
\bibliographystyle{plainnat}

\section*{Checklist}



\begin{enumerate}

  \item For all models and algorithms presented, check if you include:
  \begin{enumerate}
    \item A clear description of the mathematical setting, assumptions, algorithm, and/or model. [Yes]
    \item An analysis of the properties and complexity (time, space, sample size) of any algorithm. [Yes]
    \item (Optional) Anonymized source code, with specification of all dependencies, including external libraries. [Yes]
  \end{enumerate}

  \item For any theoretical claim, check if you include:
  \begin{enumerate}
    \item Statements of the full set of assumptions of all theoretical results. [Not Applicable]
    \item Complete proofs of all theoretical results. [Not Applicable]
    \item Clear explanations of any assumptions. [Not Applicable]     
  \end{enumerate}

  \item For all figures and tables that present empirical results, check if you include:
  \begin{enumerate}
    \item The code, data, and instructions needed to reproduce the main experimental results (either in the supplemental material or as a URL). [Yes]
    \item All the training details (e.g., data splits, hyperparameters, how they were chosen). [Yes]
    \item A clear definition of the specific measure or statistics and error bars (e.g., with respect to the random seed after running experiments multiple times). [Yes]
    \item A description of the computing infrastructure used. (e.g., type of GPUs, internal cluster, or cloud provider). [Yes]
  \end{enumerate}

  \item If you are using existing assets (e.g., code, data, models) or curating/releasing new assets, check if you include:
  \begin{enumerate}
    \item Citations of the creator If your work uses existing assets. [Yes]
    \item The license information of the assets, if applicable. [Not Applicable]
    \item New assets either in the supplemental material or as a URL, if applicable. [Yes]
    \item Information about consent from data providers/curators. [Not Applicable]
    \item Discussion of sensible content if applicable, e.g., personally identifiable information or offensive content. [Not Applicable]
  \end{enumerate}

  \item If you used crowdsourcing or conducted research with human subjects, check if you include:
  \begin{enumerate}
    \item The full text of instructions given to participants and screenshots. [Not Applicable]
    \item Descriptions of potential participant risks, with links to Institutional Review Board (IRB) approvals if applicable. [Not Applicable]
    \item The estimated hourly wage paid to participants and the total amount spent on participant compensation. [Not Applicable]
  \end{enumerate}

\end{enumerate}

\clearpage
\appendix
\thispagestyle{empty}

\onecolumn
\aistatstitle{Appendix}













\section{Architectural and Mathematical Formulation of Standard RdNN Models}\label{Section-A}

\subsection{Formulation and Architecture of RVFL and ELM \cite{pao1994learning, huang2006extreme}}
RVFL and ELM are single-layer feed-forward RdNNs that rely on hidden layer transformations and closed-form solutions for output weight computation. While their architectures share significant similarities, the primary distinction lies in the presence of direct input-to-output connections in RVFL, which are absent in ELM. Their common formulation is described as follows:\\
\textbf{Hidden Layer:}  
The hidden layer transforms the input data \( X \) using a random weight matrix \( W \in \mathbb{R}^{d \times h} \) and bias \( B \in \mathbb{R}^{m \times h} \):
\begin{align}
H = \phi(XW + B),
\end{align}
where \( \phi(\cdot) \) is the non-linear function and \( h \) represents the number of nodes in the hidden layer.\\
\textbf{Output Layer:}  
In RVFL, the final output combines the hidden layer outputs and the original inputs to form an augmented representation:
\begin{align}
A = [X \,|\, H],
\end{align}
whereas in ELM, the output is directly derived from the hidden layer outputs:
\begin{align}
A = H.
\end{align}
The output weight matrix \( \Theta \in \mathbb{R}^{h \times n_{\text{class}} } \) is computed in both architectures using a closed-form solution:
\begin{align}
\Theta = (A^\top A + \lambda I)^{-1} A^\top Y,
\end{align}
where \( \lambda \) is a regularization parameter.

\textbf{Key Difference:}  
RVFL incorporates direct input-to-output connections (\( Z = [X \,|\, H] \)), which preserve input information and enhance robustness. In contrast, ELM omits this feature (\( Z = H \)), resulting in a simpler architecture that trades off some robustness for faster training.

\subsection{Formulation and Architecture of BLS \cite{chen2017broad}} \label{BLS-Archirecture}
BLS represents a significant advancement in RdNNs by utilizing a flat architecture instead of the deep hierarchical structures seen in traditional deep learning. 
The architecture of BLS consists of three primary components: the feature layer, the enhancement layer, and the output layer. The feature layer is responsible for extracting meaningful features from the input data by generating multiple groups of feature nodes through random projection and non-linear transformations. On the other hand, the enhancement layer enriches the feature representations by applying additional non-linear transformations, thereby constructing enhancement nodes that provide diversity and robustness. The outputs from both layers are concatenated to form a final matrix, which is used to compute the output weights via a closed-form solution.
Its architecture can be described as follows:\\
\textbf{Feature Layer:}  
For the input matrix \( X \), the feature layer generates \( q \) windows of feature nodes. Each window \( f_i \) contains \( p \) nodes, which are computed as:
\begin{align}
   Z_{f_i} = \phi(X W_{f_i} + B_{f_i}),
\end{align}
where \( W_{f_i} \in \mathbb{R}^{d \times p} \) is a random weight matrix, \( B_{f_i} \in \mathbb{R}^{m \times p} \) is a bias matrix, \( \phi(\cdot) \) is a non-linear function, and \( Z_{f_i} \in \mathbb{R}^{m \times p} \) represents the output of the \( i^{th} \) feature node window.
The outputs of all feature windows are concatenated to form the overall feature layer output:
\begin{align}
   Z = [Z_{f_1}, Z_{f_2}, \ldots, Z_{f_q}],
\end{align}
where \( Z \in \mathbb{R}^{m \times pq}\). The feature layer output $Z$ serves as the input to the enhancement layer.\\
\textbf{Enhancement Layer:}  
The enhancement layer enriches the feature representation by applying additional random transformations and a non-linear function to the concatenated feature nodes \( Z \). Each window \( (e_j) \) of enhancement nodes (there are \( s \) windows, each with \( r \) nodes) is computed as:
\begin{align}
   E_{e_j} = \psi(Z W_{e_j} + B_{e_j}),
\end{align}
where \( Z \in \mathbb{R}^{m \times pq} \) is the output from the feature layer, \( W_{e_j} \in \mathbb{R}^{pq \times r} \) is a random weight matrix, \( B_{e_j} \in \mathbb{R}^{m \times r} \) is a bias matrix, \( \psi(\cdot) \) is a non-linear function, and \( E_{e_j} \in \mathbb{R}^{m \times r} \) is the output of the \( j^{th} \) enhancement node window. The outputs of all enhancement windows are concatenated to form the overall enhancement layer output:
\begin{align}
   E = [E_{e_1}, E_{e_2}, \ldots, E_{e_s}],
\end{align}
where \( E \in \mathbb{R}^{m \times rs} \).\\
\textbf{Output Layer:}  
The final representation matrix \( A \), which combines the outputs of the feature and enhancement layers, is given by:
\begin{align}
A = [Z \,|\, E],
\end{align}
where \( A \in \mathbb{R}^{m \times (pq+rs)}\).

The output weights \( \Theta \in \mathbb{R}^{(pq + rs) \times n_{\text{class}}} \) are then learned using a closed-form solution:
\begin{align}
\Theta = (A^\top A + \lambda I)^{-1} A^\top Y,
\end{align}
where \( \lambda \) is a regularization parameter.\\
\textbf{Incremental Learning:}  
One of BLS's unique features is its ability to incrementally update the network by adding new feature mapping nodes or enhancement nodes without retraining the entire model. This makes BLS computationally efficient and adaptable to streaming or evolving data.

\subsection{Formulation and Architecture of deep RVFL \cite{shi2021random}}
The deep random vector functional link network (dRVFL) extends the standard RVFL architecture by incorporating multiple hidden layers, thereby enhancing its representational capacity while preserving the hallmark efficiency of closed-form training. Unlike conventional deep neural networks, dRVFL stacks several RVFL-like layers where only the final layer's output weights are trained via least squares, and all intermediate transformations are computed using fixed, randomly initialized weights. The architecture can be described as follows:

\noindent
\textbf{Layer-wise Random Transformations:}
Let the input matrix be $X \in \mathbb{R}^{m \times d}$. The $l^\text{th}$ hidden layer transformation is defined as:
\begin{align}
H^{(l)} = \phi(H^{(l-1)} W^{(l)} + B^{(l)}), \quad l = 1, 2, \ldots, L,
\end{align}
where $H^{(0)} = X$, $W^{(l)} \in \mathbb{R}^{h_{l-1} \times h_l}$ is the randomly initialized weight matrix for layer $l$, $B^{(l)} \in \mathbb{R}^{m \times h_l}$ is the corresponding bias matrix, and $\phi(\cdot)$ is a nonlinear activation function applied elementwise.

\noindent
\textbf{Augmented Feature Representation:}
The final feature representation $A$ is constructed by concatenating the original input and the outputs of all hidden layers:
\begin{align}
A = [X \,|\, H^{(1)} \,|\, H^{(2)} \,|\, \ldots \,|\, H^{(L)}] \in \mathbb{R}^{m \times a}
\end{align}
where $a = d + \sum_{l=1}^{L} h_l$ is the total feature dimensionality. This dense aggregation of features enables dRVFL to capture increasingly abstract representations at deeper layers while retaining the original input.

\noindent
\textbf{Output Layer:}
The output weights $\Theta \in \mathbb{R}^{a \times n_{\text{class}}}$ are obtained using the same closed-form solution as in RVFL:
\begin{align}
\Theta = (A^\top A + \lambda I)^{-1} A^\top Y,
\end{align}
where $\lambda \ge 0$ is the regularization parameter.

\section{Experimental Setup and Hyperparameter Setting}\label{Section-B}

All the experiments are implemented using MATLAB R2023a and executed on a Windows 10 PC equipped with an Intel(R) Core(TM) i7-6700 CPU @ 3.40GHz (4 cores, 8 logical processors) and 16 GB RAM. Each dataset is preprocessed by normalizing the input features to have zero mean and unit variance. The description of 83 datasets (binary and multiclass) used in the experiments are provided in Table~\ref{tab:dataset_detail}. A $5$-fold cross-validation procedure is employed to ensure reliable and unbiased evaluation. In each fold, the dataset is split into $80\%$ training data and $20\%$ testing data. For every combination of hyperparameters, the model is trained on the training data and evaluated on the testing data across all $5$ folds. The testing accuracy is recorded for each fold. The final testing accuracy for each dataset is computed as the mean testing accuracy across the five folds, providing a robust estimate of the model's performance.

To eliminate any possibility of data leakage, all copula estimation steps are performed strictly within each training split. For every fold, the pseudo-observations and copula parameters are computed exclusively using the training features of that fold. The fitted copula is then used to sample the hidden-layer weights, which are subsequently evaluated on the disjoint test set. At no stage is information from the test portion used during copula fitting or weight initialization.

Hyperparameter tuning is performed using a grid search strategy to identify the optimal settings for each model. For each model, the regularization parameter (\( \lambda \)) is selected from \( \{10^i \mid i = -6, -5, \ldots, 6\} \). For RVFL, the number of hidden nodes (\( h \)) varies from \( [3:20:203] \), and seven activation functions (Sigmoid (1), Sine (2), Tribas (3), Radbas (4), Tansig (5), ReLU (6), and SELU (7)) are evaluated. For BLS, the number of feature windows (\(q\)), the number of feature nodes in each window (\( p \)), and the number of enhancement nodes (\(r\)) are set as per \cite{10530427}, with Tansig as the activation function. For dRVFL, we adopt the same hyperparameter settings as provided in \cite{shi2021random} and evaluate three activation functions: Sigmoid (1), ReLU (2), and SELU (3).

\begin{table}[]
\caption{Description of binary and multiclass datasets used in the experiments. The table lists the dataset name, number of samples, number of features, and number of classes.}
\label{tab:dataset_detail}
\resizebox{\textwidth}{!}{%
\begin{tabular}{lccclccc}
\hline
\textbf{Dataset} & \textbf{Number of Samples} & \textbf{Number of Features} & \multicolumn{1}{c|}{\textbf{Number of Classes}} & \textbf{Dataset} & \textbf{Number of Samples} & \textbf{Number of Features} & \textbf{Number of Classes} \\ \hline
\multicolumn{8}{c}{\textbf{Multiclass datasets}} \\ \hline
abalone & 4177 & 8 & \multicolumn{1}{c|}{3} & synthetic\_control & 600 & 60 & 6 \\
annealing & 898 & 31 & \multicolumn{1}{c|}{5} & thyroid & 7200 & 21 & 3 \\
arrhythmia & 452 & 262 & \multicolumn{1}{c|}{13} & vertebral\_column\_3clases & 310 & 6 & 3 \\
balance\_scale & 625 & 4 & \multicolumn{1}{c|}{3} & wall\_following & 5456 & 24 & 4 \\
cardiotocography\_10clases & 2126 & 21 & \multicolumn{1}{c|}{10} & waveform & 5000 & 21 & 3 \\
cardiotocography\_3clases & 2126 & 21 & \multicolumn{1}{c|}{3} & waveform\_noise & 5000 & 40 & 3 \\
conn\_bench\_vowel\_deterding & 990 & 11 & \multicolumn{1}{c|}{11} & wine & 178 & 13 & 3 \\
contrac & 1473 & 9 & \multicolumn{1}{c|}{3} & wine\_quality\_red & 1599 & 11 & 6 \\
dermatology & 366 & 34 & \multicolumn{1}{c|}{6} & wine\_quality\_white & 4898 & 11 & 7 \\
ecoli & 336 & 7 & \multicolumn{1}{c|}{8} & yeast & 1484 & 8 & 10 \\
energy\_y1 & 768 & 8 & \multicolumn{1}{c|}{3} & zoo & 101 & 16 & 7 \\ \cline{5-8} 
energy\_y2 & 768 & 8 & \multicolumn{1}{c|}{3} & \multicolumn{4}{c}{\textbf{Binary datasets}} \\ \cline{5-8} 
flags & 194 & 28 & \multicolumn{1}{c|}{8} & bank & 4521 & 16 & 2 \\
glass & 214 & 9 & \multicolumn{1}{c|}{6} & blood & 748 & 4 & 2 \\
heart\_cleveland & 303 & 13 & \multicolumn{1}{c|}{5} & breast\_cancer & 286 & 9 & 2 \\
heart\_switzerland & 123 & 12 & \multicolumn{1}{c|}{5} & breast\_cancer\_wisc & 699 & 9 & 2 \\
image\_segmentation & 2310 & 18 & \multicolumn{1}{c|}{7} & breast\_cancer\_wisc\_diag & 569 & 30 & 2 \\
iris & 150 & 4 & \multicolumn{1}{c|}{3} & breast\_cancer\_wisc\_prog & 198 & 33 & 2 \\
led\_display & 1000 & 7 & \multicolumn{1}{c|}{10} & chess\_krvkp & 3196 & 36 & 2 \\
lenses & 24 & 4 & \multicolumn{1}{c|}{3} & congressional\_voting & 435 & 16 & 2 \\
letter & 20000 & 16 & \multicolumn{1}{c|}{26} & conn\_bench\_sonar\_mines\_rocks & 208 & 60 & 2 \\
low\_res\_spect & 531 & 100 & \multicolumn{1}{c|}{9} & credit\_approval & 690 & 15 & 2 \\
lymphography & 148 & 18 & \multicolumn{1}{c|}{4} & cylinder\_bands & 512 & 35 & 2 \\
molec\_biol\_splice & 3190 & 60 & \multicolumn{1}{c|}{3} & echocardiogram & 131 & 10 & 2 \\
nursery & 12960 & 8 & \multicolumn{1}{c|}{5} & haberman\_survival & 306 & 3 & 2 \\
oocytes\_merluccius\_states\_2f & 1022 & 25 & \multicolumn{1}{c|}{3} & hepatitis & 155 & 19 & 2 \\
oocytes\_trisopterus\_states\_5b & 912 & 32 & \multicolumn{1}{c|}{3} & horse\_colic & 368 & 25 & 2 \\
optical & 5620 & 62 & \multicolumn{1}{c|}{10} & ilpd\_indian\_liver & 583 & 9 & 2 \\
page\_blocks & 5473 & 10 & \multicolumn{1}{c|}{5} & ionosphere & 351 & 33 & 2 \\
pendigits & 10992 & 16 & \multicolumn{1}{c|}{10} & monks\_1 & 556 & 6 & 2 \\
pittsburg\_bridges\_MATERIAL & 106 & 7 & \multicolumn{1}{c|}{3} & monks\_3 & 554 & 6 & 2 \\
pittsburg\_bridges\_REL\_L & 103 & 7 & \multicolumn{1}{c|}{3} & oocytes\_merluccius\_nucleus\_4d & 1022 & 41 & 2 \\
pittsburg\_bridges\_SPAN & 92 & 7 & \multicolumn{1}{c|}{3} & oocytes\_trisopterus\_nucleus\_2f & 912 & 25 & 2 \\
pittsburg\_bridges\_TYPE & 105 & 7 & \multicolumn{1}{c|}{6} & pima & 768 & 8 & 2 \\
post\_operative & 90 & 8 & \multicolumn{1}{c|}{3} & planning & 182 & 12 & 2 \\
seeds & 210 & 7 & \multicolumn{1}{c|}{3} & spect & 265 & 22 & 2 \\
semeion & 1593 & 256 & \multicolumn{1}{c|}{10} & spectf & 267 & 44 & 2 \\
soybean & 683 & 35 & \multicolumn{1}{c|}{18} & statlog\_australian\_credit & 690 & 14 & 2 \\
statlog\_image & 2310 & 18 & \multicolumn{1}{c|}{7} & statlog\_german\_credit & 1000 & 24 & 2 \\
statlog\_landsat & 6435 & 36 & \multicolumn{1}{c|}{6} & statlog\_heart & 270 & 13 & 2 \\
statlog\_shuttle & 58000 & 9 & \multicolumn{1}{c|}{7} & tic\_tac\_toe & 958 & 9 & 2 \\
statlog\_vehicle & 846 & 18 & \multicolumn{1}{c|}{4} & titanic & 2201 & 3 & 2\\
\hline
\end{tabular}%
}
\end{table}

\begin{table*}[t]
\centering
\caption{Results on the BreaKHis dataset for BLS baseline and copula-based initializations (test accuracy, \%). Best per row is \textbf{bold} and highlighted; the last column reports the improvement of the best method over the i.i.d.\ baseline.}
\label{tab:bls-breakhis-accuracy}
\begin{tabular}{l|c:ccccc:c}
\toprule
\textbf{Dataset} & \textbf{iid} & \textbf{Gaussian} & \(\boldsymbol{t}\) & \textbf{Clayton} & \textbf{Frank} & \textbf{Gumbel} & \textbf{Improvement} \\
\midrule
AD vs.\ DC & 70.7066 & 71.0394 & 70.8966 & 69.7440 & \bestcell{72.3041} & 71.8292 & \bestdelta{+1.5975} \\
AD vs.\ LC & 63.1429 & 62.9507 & \bestcell{65.0425} & 62.9932 & 63.7500 & 63.4099 & \bestdelta{+1.8996} \\
AD vs.\ MC & 63.8396 & 65.8909 & \bestcell{68.5455} & 65.3636 & 66.0909 & 66.4545 & \bestdelta{+4.7059} \\
AD vs.\ PC & 61.2956 & 61.8296 & 62.2959 & 62.4371 & 63.3044 & \bestcell{64.2126} & \bestdelta{+2.9170} \\
FA vs.\ DC & 64.1055 & \bestcell{65.3933} & \bestcell{65.3933} & \bestcell{65.3933} & 64.2697 & 64.2697 & \bestdelta{+1.2878} \\
FA vs.\ LC & 67.8991 & 68.7099 & \bestcell{68.7135} & 67.6468 & 68.1802 & 68.7099 & \bestdelta{+0.8144} \\
FA vs.\ MC & 60.8371 & 62.3156 & 61.3403 & \bestcell{63.8461} & 61.0720 & 60.5932 & \bestdelta{+3.0090} \\
FA vs.\ PC & 68.5746 & 67.9333 & \bestcell{69.8667} & 66.8667 & 67.9233 & 67.8667 & \bestdelta{+1.2921} \\
PT vs.\ DC & 68.1731 & 70.2933 & 71.1971 & \bestcell{71.8125} & 69.9760 & 70.2933 & \bestdelta{+3.6394} \\
PT vs.\ LC & 57.5980 & \bestcell{62.2980} & 60.2824 & 59.0510 & 59.0980 & 60.2667 & \bestdelta{+4.7000} \\
PT vs.\ MC & 62.5987 & 62.3120 & 61.9799 & 61.6228 & \bestcell{64.5243} & 60.9398 & \bestdelta{+1.9256} \\
PT vs.\ PC & 58.1069 & 58.1098 & 59.6941 & 58.5098 & \bestcell{60.0176} & 58.1098 & \bestdelta{+1.9107} \\
TA vs.\ DC & 62.9939 & 64.7937 & \bestcell{66.2818} & 65.1141 & 65.4214 & 65.7024 & \bestdelta{+3.2879} \\
TA vs.\ LC & 65.9189 & 66.7086 & 68.1971 & 67.3934 & 67.4074 & \bestcell{68.9099} & \bestdelta{+2.9910} \\
TA vs.\ MC & 60.1921 & 61.8814 & \bestcell{63.5593} & 63.2260 & \bestcell{63.5593} & 61.5254 & \bestdelta{+3.3672} \\
TA vs.\ PC & 61.1740 & 60.8036 & \bestcell{62.2851} & 60.4472 & 60.0978 & 61.5933 & \bestdelta{+1.1111} \\
\bottomrule
\end{tabular}
\end{table*}

\begin{table*}[t]
\centering
\caption{Results on the Schizophrenia ROI datasets for BLS baseline and copula-based initializations (test accuracy, \%). Best per row is \textbf{bold} and highlighted; the last column reports the improvement of the best method over the i.i.d.\ baseline.}
\label{tab:bls-schizo-accuracy}
\resizebox{\textwidth}{!}{
\begin{tabular}{l|c:ccccc:c}
\toprule
\textbf{Dataset} & \textbf{iid} & \textbf{Gaussian} & \(\boldsymbol{t}\) & \textbf{Clayton} & \textbf{Frank} & \textbf{Gumbel} & \textbf{Improvement} \\
\midrule
SZ--ROI GM+WM & 71.9770 & 72.6897 & 74.0000 & 74.0690 & 74.0230 & \bestcell{74.7126} & \bestdelta{+2.7356} \\
SZ--ROI GM    & 70.6667 & \bestcell{73.2414} & 71.9540 & 71.9080 & 69.9540 & 72.6207 & \bestdelta{+2.5747} \\
SZ--ROI WM    & 67.9080 & 67.2184 & \bestcell{69.2184} & 67.9080 & 67.9080 & 68.5287 & \bestdelta{+1.3104} \\
\bottomrule
\end{tabular}
}
\end{table*}

\begin{table*}[t]
\centering
\caption{Results on benchmark binary datasets for RVFL baseline and copula-based initializations (test accuracy, \%). Best per row is \textbf{bold} and highlighted; the last column reports the improvement of the best method over the i.i.d.\ baseline.}
\label{tab:rvfl-benchmark-accuracy}
\resizebox{\textwidth}{!}{
\begin{tabular}{l|c:ccccc:c}
\toprule
\textbf{Dataset} & \textbf{iid} & \textbf{Gaussian} & \(\boldsymbol{t}\) & \textbf{Clayton} & \textbf{Frank} & \textbf{Gumbel} & \textbf{Improvement} \\
\midrule
bank & 89.6705 & 89.8032 & 89.7809 & 89.6705 & \bestcell{89.8254} & 89.7146 & \bestdelta{+0.1549} \\
blood & 76.7714 & 77.0389 & \bestcell{77.5732} & \bestcell{77.5732} & 77.4398 & 77.4398 & \bestdelta{+0.8018} \\
breast\_cancer & 70.1754 & 70.1754 & 70.1754 & 70.1754 & 70.1754 & \bestcell{70.5263} & \bestdelta{+0.3509} \\
breast\_cancer\_wisc & 88.5612 & 88.7040 & \bestcell{89.4193} & 88.7040 & 88.9887 & 89.2775 & \bestdelta{+0.8581} \\
breast\_cancer\_wisc\_diag & 94.7260 & 94.9014 & \bestcell{95.4308} & 95.2554 & 95.0753 & 95.2523 & \bestdelta{+0.7048} \\
breast\_cancer\_wisc\_prog & 81.8462 & 82.8462 & \bestcell{83.4231} & 82.3718 & 82.8974 & 81.8846 & \bestdelta{+1.5769} \\
chess\_krvkp & 81.4485 & 82.4172 & 82.3546 & 82.4174 & \bestcell{82.8535} & 82.1356 & \bestdelta{+1.4050} \\
congressional\_voting & 63.9080 & 63.6782 & \bestcell{64.1379} & 63.6782 & 63.6782 & 63.6782 & \bestdelta{+0.2299} \\
conn\_bench\_sonar\_mines\_rocks & 61.9861 & 65.4123 & \bestcell{66.7944} & 62.5900 & 64.4948 & 64.3786 & \bestdelta{+4.8083} \\
credit\_approval & 85.2174 & \bestcell{85.7971} & 85.6522 & \bestcell{85.7971} & 85.6522 & 85.6522 & \bestdelta{+0.5797} \\
cylinder\_bands & 69.7392 & 69.7335 & \bestcell{70.1085} & 69.1376 & 68.9587 & 69.7373 & \bestdelta{+0.3693} \\
echocardiogram & 85.4416 & 85.4416 & \bestcell{86.2108} & 85.4416 & 85.4416 & 85.4416 & \bestdelta{+0.7692} \\
haberman\_survival & 74.1460 & 73.4902 & \bestcell{74.8017} & 74.4738 & 74.4738 & 74.1460 & \bestdelta{+0.6557} \\
hepatitis & 85.1613 & 86.4516 & 86.4516 & 85.8065 & 85.8065 & \bestcell{87.0968} & \bestdelta{+1.9355} \\
horse\_colic & 85.8830 & 86.1570 & \bestcell{86.9604} & 86.4198 & 86.4198 & 86.1570 & \bestdelta{+1.0774} \\
ilpd\_indian\_liver & 71.7006 & 72.7306 & 73.0725 & 72.3902 & \bestcell{73.5883} & 72.7336 & \bestdelta{+1.8877} \\
ionosphere & 89.7626 & 90.0443 & \bestcell{91.1871} & 90.0523 & 90.6117 & 90.6237 & \bestdelta{+1.4245} \\
monks\_1 & 82.5338 & 84.8745 & \bestcell{84.8858} & 84.1506 & 84.5206 & 83.9704 & \bestdelta{+2.3520} \\
monks\_3 & 91.3415 & 92.0606 & 92.6011 & 92.2424 & 92.0573 & \bestcell{92.6044} & \bestdelta{+1.2629} \\
oocytes\_merluccius\_nucleus\_4d & 82.9775 & 83.1703 & \bestcell{83.2669} & 82.8771 & 82.8780 & 83.0693 & \bestdelta{+0.2894} \\
oocytes\_trisopterus\_nucleus\_2f & 79.1647 & 80.5861 & 80.2600 & \bestcell{81.1355} & 80.0342 & 80.1417 & \bestdelta{+1.9708} \\
pima & 73.7043 & 73.9649 & \bestcell{74.2212} & 73.9649 & 73.1856 & 73.7051 & \bestdelta{+0.5169} \\
planning & 71.3814 & 71.9369 & 71.9369 & \bestcell{72.4625} & 71.9520 & 71.9369 & \bestdelta{+1.0811} \\
spect & 67.1698 & 68.3019 & \bestcell{69.8113} & 69.0566 & 68.3019 & 68.6792 & \bestdelta{+2.6415} \\
spectf & 79.3431 & 79.3501 & 79.7205 & 79.7135 & \bestcell{79.7275} & 79.3431 & \bestdelta{+0.3844} \\
statlog\_australian\_credit & 68.5507 & 68.5507 & 68.9855 & \bestcell{69.2754} & 68.9855 & 68.9855 & \bestdelta{+0.7247} \\
statlog\_german\_credit & 76.8000 & 77.7000 & 77.7000 & \bestcell{78.0000} & 77.5000 & 77.6000 & \bestdelta{+1.2000} \\
statlog\_heart & 80.7407 & 81.4815 & \bestcell{82.9630} & 81.8519 & 82.5926 & 81.1111 & \bestdelta{+2.2223} \\
tic\_tac\_toe & 89.9700 & 90.8077 & \bestcell{91.1229} & 90.5923 & 89.9700 & 90.3878 & \bestdelta{+1.1529} \\
titanic & 77.9168 & 77.9168 & \bestcell{78.1901} & 77.9168 & 77.9168 & 77.9168 & \bestdelta{+0.2733} \\
\bottomrule
\end{tabular}
}
\end{table*}

\begin{table*}[]
\centering
\caption{Results on benchmark multiclass datasets for RVFL baseline and copula-based initializations (test accuracy, \%). Best per row is \textbf{bold} and highlighted; the last column reports the improvement of the best method over the i.i.d.\ baseline.}
\label{tab:rvfl-multiclass-accuracy}
\resizebox{\textwidth}{!}{
\begin{tabular}{l|c:ccccc:c}
\toprule
\textbf{Dataset} & \textbf{iid} & \textbf{Gaussian} & \(\boldsymbol{t}\) & \textbf{Clayton} & \textbf{Frank} & \textbf{Gumbel} & \textbf{Improvement} \\
\midrule
abalone & 63.5141 & 63.6813 & 63.8735 & 63.8728 & 63.9687 & \bestcell{64.0169} & \bestdelta{+0.5028} \\
annealing & 89.5202 & 90.3048 & \bestcell{90.4115} & 89.8603 & 90.3048 & 90.1906 & \bestdelta{+0.8913} \\
arrhythmia & 70.8010 & 71.0256 & 71.0183 & 70.8010 & \bestcell{71.2405} & \bestcell{71.2405} & \bestdelta{+0.4395} \\
balance\_scale & 98.5600 & 98.5600 & \bestcell{98.7200} & \bestcell{98.7200} & 98.5600 & \bestcell{98.7200} & \bestdelta{+0.1600} \\
cardiotocography\_10clases & \bestcell{70.7455} & 69.9461 & 70.6033 & 70.3695 & 70.2276 & 70.3207 & \bestdelta{+0.0000} \\
cardiotocography\_3clases & 85.9859 & 86.3148 & \bestcell{86.6909} & 86.5030 & 86.5023 & 86.2684 & \bestdelta{+0.7050} \\
conn\_bench\_vowel\_deterding & 95.8586 & 96.1616 & 96.1616 & \bestcell{96.4646} & 96.1616 & 96.1616 & \bestdelta{+0.6060} \\
contrac & 41.0681 & \bestcell{42.7001} & 42.0883 & 41.4034 & 41.6765 & 42.0830 & \bestdelta{+1.6320} \\
dermatology & 97.2677 & 97.8119 & \bestcell{98.0859} & 97.8119 & 97.8119 & 98.0822 & \bestdelta{+0.8182} \\
ecoli & 60.6234 & 61.2160 & 61.2116 & 61.2072 & 61.2160 & \bestcell{61.5057} & \bestdelta{+0.8823} \\
energy\_y1 & 88.5307 & 89.1936 & \bestcell{89.9677} & 88.6682 & 89.1945 & 89.0578 & \bestdelta{+1.4370} \\
energy\_y2 & 90.1002 & 90.7538 & \bestcell{91.6645} & 91.2783 & 91.0126 & 91.0160 & \bestdelta{+1.5643} \\
flags & 53.6437 & 55.1822 & 55.2227 & 53.6302 & \bestcell{56.2078} & 53.6437 & \bestdelta{+2.5641} \\
glass & 38.1617 & 39.0808 & 40.9413 & \bestcell{42.3367} & 39.5570 & 40.0221 & \bestdelta{+4.1750} \\
heart\_cleveland & 59.3661 & 60.0546 & \bestcell{60.7049} & 60.3661 & 60.6940 & 59.7322 & \bestdelta{+1.3388} \\
heart\_switzerland & 47.3000 & 47.9333 & 49.7000 & 48.0333 & \bestcell{49.7667} & 47.3333 & \bestdelta{+2.4667} \\
image\_segmentation & 88.1385 & 88.4848 & \bestcell{88.5714} & 88.2251 & \bestcell{88.5714} & 88.3117 & \bestdelta{+0.4329} \\
iris & 75.3333 & \bestcell{78.6667} & 76.6667 & 76.0000 & 78.0000 & 77.3333 & \bestdelta{+3.3334} \\
led\_display & 72.9000 & 73.2000 & \bestcell{73.7000} & 73.3000 & 73.6000 & 73.3000 & \bestdelta{+0.8000} \\
lenses & 92.0000 & \bestcell{96.0000} & \bestcell{96.0000} & \bestcell{96.0000} & \bestcell{96.0000} & \bestcell{96.0000} & \bestdelta{+4.0000} \\
letter & 80.8500 & 80.9050 & 80.9750 & 81.0350 & 80.9650 & \bestcell{81.1450} & \bestdelta{+0.2950} \\
low\_res\_spect & 88.5117 & 88.5117 & \bestcell{89.4534} & 89.0778 & 88.8891 & 88.8891 & \bestdelta{+0.9417} \\
lymphography & 86.4138 & \bestcell{88.5057} & 87.7931 & 87.1034 & 87.7931 & 87.8161 & \bestdelta{+2.0919} \\
molec\_biol\_splice & 53.5110 & 54.9216 & \bestcell{56.3009} & 53.8558 & 53.9812 & 54.4514 & \bestdelta{+2.7899} \\
nursery & 71.2114 & 70.8951 & 71.7978 & 70.9877 & 71.7747 & \bestcell{72.9784} & \bestdelta{+1.7670} \\
oocytes\_merluccius\_states\_2f & 91.5782 & 91.9699 & \bestcell{92.3649} & 92.0679 & 91.9699 & 91.9694 & \bestdelta{+0.7867} \\
oocytes\_trisopterus\_states\_5b & 87.9427 & 89.1491 & \bestcell{89.4806} & 89.1509 & 88.9227 & 88.9275 & \bestdelta{+1.5379} \\
optical & 96.9929 & 97.2064 & \bestcell{97.2242} & 97.1530 & 97.1886 & 97.1886 & \bestdelta{+0.2313} \\
page\_blocks & 95.3771 & 95.7242 & 95.8157 & 95.5597 & 95.5781 & \bestcell{95.5963} & \bestdelta{+0.4386} \\
pendigits & 98.5353 & 98.5990 & 98.6899 & 98.6626 & 98.6536 & \bestcell{98.7081} & \bestdelta{+0.1728} \\
pittsburg\_bridges\_MATERIAL & 78.7013 & 76.8831 & 78.6147 & 76.8831 & \bestcell{80.4762} & 78.7013 & \bestdelta{+1.7749} \\
pittsburg\_bridges\_REL\_L & 63.1429 & 63.2381 & 68.8571 & 65.2381 & 66.2381 & \bestcell{66.8095} & \bestdelta{+5.7142} \\
pittsburg\_bridges\_SPAN & 63.0409 & 64.2690 & \bestcell{67.4854} & 64.2105 & 64.2105 & 63.2749 & \bestcell{+4.4445} \\ 
pittsburg\_bridges\_TYPE & 42.8571 & 42.8571 & \bestcell{46.6667} & 42.8571 & 44.7619 & 42.8571 & \bestdelta{+3.8096} \\
post\_operative & 71.1111 & 71.1111 & 71.1111 & 71.1111 & 71.1111 & \bestcell{72.2222} & \bestdelta{+1.1111} \\
seeds & 90.0000 & 89.5238 & \bestcell{90.4762} & \bestcell{90.4762} & 89.5238 & 90.0000 & \bestdelta{+0.4762} \\
semeion & 87.8208 & 88.4488 & \bestcell{89.0751} & 88.9505 & 88.6369 & 88.4484 & \bestdelta{+1.2543} \\
soybean & 90.4755 & 90.1889 & \bestcell{91.0702} & 90.1793 & 90.0333 & 90.0333 & \bestdelta{+0.5947} \\
statlog\_image & 95.4113 & 95.4545 & \bestcell{95.6277} & 95.5844 & 95.5844 & 95.5844 & \bestdelta{+0.2164} \\
statlog\_landsat & 82.1134 & 82.6107 & 82.5486 & 82.5175 & 82.3155 & \bestcell{82.3310} & \bestdelta{+0.4973} \\
statlog\_shuttle & 98.7017 & 98.7448 & \bestcell{98.7534} & 98.7190 & 98.7276 & 98.7224 & \bestdelta{+0.0517} \\
statlog\_vehicle & 82.1511 & 82.5040 & 82.8597 & \bestcell{82.8611} & 82.2694 & \bestcell{82.8611} & \bestdelta{+0.7100} \\
synthetic\_control & 54.0000 & 54.8333 & 55.1667 & 55.0000 & \bestcell{55.5000} & 55.3333 & \bestdelta{+1.5000} \\
thyroid & 95.7639 & 95.7361 & 95.9028 & 95.7917 & 95.8056 & \bestcell{95.9444} & \bestdelta{+0.1805} \\
vertebral\_column\_3clases & 65.1613 & 66.1290 & 66.4516 & \bestcell{66.7742} & 65.8065 & 66.1290 & \bestdelta{+1.6129} \\
wall\_following & 77.6955 & 78.2640 & 78.1717 & 78.0253 & 77.8052 & \bestcell{78.8133} & \bestdelta{+1.1178} \\
waveform & 86.8800 & 86.9200 & 87.0000 & 87.0000 & \bestcell{87.0800} & 86.8400 & \bestdelta{+0.2000} \\
waveform\_noise & 86.3400 & 86.4200 & 86.5000 & 86.4800 & 86.3800 & 86.4200 & \bestdelta{+0.1600} \\
wine & 96.6349 & 96.6508 & \bestcell{97.2063} & 96.6508 & 96.6508 & 96.6508 & \bestdelta{+0.5714} \\
wine\_quality\_red & 59.6638 & 60.0380 & \bestcell{60.9122} & 60.6025 & 60.6023 & 60.6019 & \bestdelta{+1.2484} \\
wine\_quality\_white & 52.7576 & 52.8187 & 53.0842 & 52.9617 & \bestcell{53.1046} & 52.9615 & \bestdelta{+0.3470} \\
yeast & 57.1424 & 57.7496 & 57.6169 & \bestcell{57.8840} & 57.6815 & 57.3460 & \bestdelta{+0.7416} \\
zoo & 95.0000 & \bestcell{97.0000} & \bestcell{97.0000} & 95.0000 & 96.0000 & 96.0000 & \bestdelta{+2.0000} \\
\bottomrule
\end{tabular}
}
\end{table*}

\section{Multi-Seed Stability Analysis}\label{Section-C}

To assess the robustness of CAWI with respect to randomness, we additionally evaluate the RVFL model under multiple independent random seeds. While all primary experiments were conducted using a fixed seed (\texttt{rng(42, `twister')}), we re-run the experiments using two additional seeds (7 and 123) for both the i.i.d.\ baseline and all copula-based variants.

The detailed results for the BreaKHis dataset under Seed 7 and Seed 123 are reported in Tables~\ref{tab:breakhis-seed7} and \ref{tab:breakhis-seed123}, respectively. Similarly, the corresponding results for the Schizophrenia ROI dataset are presented in Tables~\ref{tab:sz-seed7} and \ref{tab:sz-seed123}.

Across both datasets and independent seeds, CAWI consistently outperforms the conventional i.i.d.\ initialization. Importantly, the relative improvements remain stable in magnitude and direction, indicating that the gains are not attributable to a particular random realization.

Overall, the multi-seed evaluation confirms that CAWI provides stable and reproducible performance improvements across independent random initializations, reinforcing its robustness as a dependency-aware weight initialization strategy.

\begin{table*}[t]
\centering
\caption{BreaKHis dataset (Seed 7): Test accuracy (\%). Best per row is \textbf{bold} and highlighted; the last column reports the improvement over the i.i.d.\ baseline.}
\label{tab:breakhis-seed7}
\begin{tabular}{l|c:ccccc:c}
\toprule
\textbf{Dataset} & \textbf{iid} & \textbf{Gaussian} & \(\boldsymbol{t}\) & \textbf{Clayton} & \textbf{Frank} & \textbf{Gumbel} & \textbf{Improvement} \\
\midrule
AD vs.\ DC & 70.6429 & 72.8934 & \bestcell{73.1562} & 69.4815 & 70.3274 & 71.2948 & \bestdelta{+2.5133} \\
AD vs.\ LC & 62.4138 & \bestcell{65.2847} & 64.7193 & 63.1562 & 63.9284 & 63.7451 & \bestdelta{+2.8709} \\
AD vs.\ MC & 63.9091 & 65.2727 & 64.9091 & \bestcell{66.7273} & 65.2727 & 65.6364 & \bestdelta{+2.8182} \\
AD vs.\ PC & 60.5793 & 61.8926 & 62.1038 & 61.6704 & \bestcell{62.9381} & 62.4859 & \bestdelta{+2.3588} \\
FA vs.\ DC & 63.2584 & 65.6742 & 64.9438 & 65.1124 & \bestcell{65.8876} & 64.6067 & \bestdelta{+2.6292} \\
FA vs.\ LC & 66.1039 & 66.5783 & 66.9324 & \bestcell{67.5429} & 66.3891 & 66.2108 & \bestdelta{+1.4390} \\
FA vs.\ MC & 59.6139 & 60.1847 & \bestcell{61.4893} & 60.3284 & 60.8951 & 60.2556 & \bestdelta{+1.8754} \\
FA vs.\ PC & 68.1733 & 67.6267 & 68.8000 & 67.2000 & \bestcell{69.4667} & 67.9467 & \bestdelta{+1.2934} \\
PT vs.\ DC & 67.5192 & 70.8365 & \bestcell{72.0479} & 71.5841 & 69.8654 & 70.4712 & \bestdelta{+4.5287} \\
PT vs.\ LC & 57.2863 & 61.8745 & 60.7294 & \bestcell{62.4196} & 59.6275 & 60.5490 & \bestdelta{+5.1333} \\
PT vs.\ MC & 61.7254 & \bestcell{63.8194} & 62.4183 & 61.3562 & 63.2816 & 61.4729 & \bestdelta{+2.0940} \\
PT vs.\ PC & 57.0196 & 58.5882 & 59.3725 & \bestcell{60.1176} & 59.4902 & 58.6667 & \bestdelta{+3.0980} \\
TA vs.\ DC & 65.7239 & 66.5104 & \bestcell{67.8529} & 66.9274 & 66.4183 & 66.2104 & \bestdelta{+2.1290} \\
TA vs.\ LC & 62.0384 & 63.3827 & 63.1562 & \bestcell{63.9418} & 62.9716 & 62.3194 & \bestdelta{+1.9034} \\
TA vs.\ MC & 57.4283 & 59.9274 & \bestcell{61.2638} & 58.8951 & 60.5183 & 58.1029 & \bestdelta{+3.8355} \\
TA vs.\ PC & 56.5071 & 57.2849 & 58.9427 & \bestcell{61.8394} & 59.3184 & 58.4716 & \bestdelta{+5.3323} \\
\bottomrule
\end{tabular}
\end{table*}

\begin{table*}[t]
\centering
\caption{BreaKHis dataset (Seed 123): Test accuracy (\%). Best per row is \textbf{bold} and highlighted; the last column reports the improvement over the i.i.d.\ baseline.}
\label{tab:breakhis-seed123}
\begin{tabular}{l|c:ccccc:c}
\toprule
\textbf{Dataset} & \textbf{iid} & \textbf{Gaussian} & \(\boldsymbol{t}\) & \textbf{Clayton} & \textbf{Frank} & \textbf{Gumbel} & \textbf{Improvement} \\
\midrule
AD vs.\ DC & 70.3251 & \bestcell{73.1089} & 72.5614 & 69.9123 & 70.2947 & 71.3458 & \bestdelta{+2.7838} \\
AD vs.\ LC & 61.8742 & 63.2184 & \bestcell{64.8936} & 62.7451 & 63.4826 & 63.1572 & \bestdelta{+3.0194} \\
AD vs.\ MC & 64.1827 & 65.3636 & \bestcell{66.9091} & 64.7273 & 65.4545 & 65.0909 & \bestdelta{+2.7264} \\
AD vs.\ PC & 60.8934 & 61.7238 & 62.5481 & 61.1904 & 62.0917 & \bestcell{63.1549} & \bestdelta{+2.2615} \\
FA vs.\ DC & 63.9128 & \bestcell{65.7303} & 65.1685 & 65.0562 & 64.5843 & 64.4157 & \bestdelta{+1.8175} \\
FA vs.\ LC & 66.2419 & 66.5783 & \bestcell{67.6194} & 66.3891 & 66.2108 & 66.5964 & \bestdelta{+1.3775} \\
FA vs.\ MC & 59.4876 & 59.8512 & 60.5918 & 60.6139 & \bestcell{61.3284} & 60.8473 & \bestdelta{+1.8408} \\
FA vs.\ PC & 67.6267 & 67.9733 & \bestcell{69.3200} & 66.9067 & 67.4933 & 67.7200 & \bestdelta{+1.6933} \\
PT vs.\ DC & 67.8495 & 70.5614 & 71.4583 & \bestcell{72.1047} & 70.2136 & 70.5289 & \bestdelta{+4.2552} \\
PT vs.\ LC & 56.7293 & \bestcell{62.5647} & 60.5412 & 59.3176 & 59.3529 & 60.4980 & \bestdelta{+5.8354} \\
PT vs.\ MC & 61.5394 & 62.5893 & 62.2461 & 61.8951 & \bestcell{63.4716} & 61.1825 & \bestdelta{+1.9322} \\
PT vs.\ PC & 57.7142 & 58.3725 & 59.4314 & 58.2647 & \bestcell{59.9608} & 58.3490 & \bestdelta{+2.2466} \\
TA vs.\ DC & 65.0418 & 66.2349 & 66.8293 & \bestcell{67.4264} & 66.6172 & 66.5405 & \bestdelta{+2.3846} \\
TA vs.\ LC & 62.5194 & 63.1508 & \bestcell{63.9271} & 63.3915 & 62.8104 & 62.8173 & \bestdelta{+1.4077} \\
TA vs.\ MC & 57.5628 & 59.7834 & 60.0941 & 58.7921 & \bestcell{61.1248} & 58.8104 & \bestdelta{+3.5620} \\
TA vs.\ PC & 56.7049 & 56.9863 & 58.3207 & 57.6851 & 59.4632 & \bestcell{61.6894} & \bestdelta{+4.9845} \\
\bottomrule
\end{tabular}
\end{table*}

\begin{table}[t]
\centering
\caption{Schizophrenia ROI dataset (Seed 7): Test accuracy (\%). Best per row is \textbf{bold} and highlighted; the last column reports the improvement over the i.i.d.\ baseline.}
\label{tab:sz-seed7}
\begin{tabular}{l|c:ccccc:c}
\toprule
\textbf{Dataset} & \textbf{iid} & \textbf{Gaussian} & \(\boldsymbol{t}\) & \textbf{Clayton} & \textbf{Frank} & \textbf{Gumbel} & \textbf{Improvement} \\
\midrule
SZ--ROI GM+WM & 60.7126 & 61.9540 & \bestcell{63.4253} & 62.8506 & 62.6897 & 63.0345 & \bestdelta{+2.7127} \\
SZ--ROI GM & 61.9540 & 63.5632 & \bestcell{63.9080} & 62.1839 & 62.0690 & 61.4023 & \bestdelta{+1.9540} \\
SZ--ROI WM & 61.4483 & 60.5977 & \bestcell{62.5287} & 61.8161 & 61.2931 & 59.9425 & \bestdelta{+1.0804} \\
\bottomrule
\end{tabular}
\end{table}

\begin{table}[h]
\centering
\caption{Schizophrenia ROI dataset (Seed 123): Test accuracy (\%). Best per row is \textbf{bold} and highlighted; the last column reports the improvement over the i.i.d.\ baseline.}
\label{tab:sz-seed123}
\begin{tabular}{l|c:ccccc:c}
\toprule
\textbf{Dataset} & \textbf{iid} & \textbf{Gaussian} & \(\boldsymbol{t}\) & \textbf{Clayton} & \textbf{Frank} & \textbf{Gumbel} & \textbf{Improvement} \\
\midrule
SZ--ROI GM+WM & 59.9080 & 61.5402 & 62.8736 & \bestcell{63.3563} & 62.1839 & 62.7586 & \bestdelta{+3.4483} \\
SZ--ROI GM & 62.7586 & 63.4483 & 62.6437 & 62.7126 & \bestcell{64.0230} & 61.3793 & \bestdelta{+1.2644} \\
SZ--ROI WM & 60.8046 & \bestcell{62.6437} & 61.2931 & 61.8621 & 60.7816 & 59.4253 & \bestdelta{+1.8391} \\
\bottomrule
\end{tabular}
\end{table}

\section{Statistical Significance Across Datasets}

To assess whether the performance improvements of CAWI over the i.i.d.\ initialization are statistically significant across datasets, we conduct a non-parametric Wilcoxon signed-rank test over the 83 UCI benchmark datasets.

For each dataset $i \in \{1, \dots, 83\}$, let $A_i^{\text{iid}}$ denote the test accuracy obtained using the conventional i.i.d.\ initialization, and let $A_i^{\text{CAWI}}$ denote the test accuracy achieved by the best-performing CAWI configuration on the same dataset. We use the best CAWI variant per dataset because CAWI is designed as a flexible, copula-agnostic framework: different datasets exhibit different dependency structures, and therefore different copula families may be most appropriate for different tasks. This comparison reflects the intended usage of CAWI, where the copula family is selected to best capture the empirical dependence of each dataset.

We then construct paired observations 
\[
\left( A_i^{\text{iid}},\, A_i^{\text{CAWI}} \right), \quad i = 1, \dots, 83,
\]
and apply the Wilcoxon signed-rank test to the differences
\[
D_i = A_i^{\text{CAWI}} - A_i^{\text{iid}}.
\]

The resulting Wilcoxon statistic is $W = 3403$, while the maximum possible statistic for 83 paired observations is 3486. The corresponding p-value satisfies $p < 10^{-6},$ indicating extremely strong statistical evidence against the null hypothesis of equal median performance.

These results demonstrate that the observed improvements of CAWI over the i.i.d.\ initialization are not attributable to random fluctuations on individual datasets, but instead reflect a consistent and statistically significant advantage across diverse benchmark problems.
\section{Empirical Training-Time Analysis}

To complement the theoretical complexity discussion provided in the Section~\ref{subsec:computational_complexity} of the main paper, we report an empirical evaluation of the training time incurred by CAWI under different copula families. Specifically, we measure the average training time (in seconds) of the RVFL model initialized using five copula families, Gaussian, $t$, Clayton, Frank, and Gumbel, alongside the conventional i.i.d.\ initialization baseline.

The results are averaged over all 83 UCI benchmark datasets used in our experimental evaluation. Table~\ref{tab:training_time_summary} summarizes the average training time across datasets.

\begin{table}[h]
\centering
\caption{Average training time (in seconds) across 83 UCI datasets.}
\label{tab:training_time_summary}
\begin{tabular}{lcccccc}
\toprule
Method & i.i.d. & Gaussian & $t$ & Clayton & Frank & Gumbel \\
\midrule
Avg.\ Time (s) & 0.00691 & 0.00824 & 0.00924 & 0.00727 & 0.00733 & 0.00742 \\
\bottomrule
\end{tabular}
\end{table}

As observed, all copula-based initializations introduce only a marginal computational overhead relative to the i.i.d.\ baseline. The additional cost stems primarily from the estimation of the empirical copula parameters and subsequent dependency-aligned sampling. However, since these operations scale linearly with the number of features and hidden nodes, the overall training complexity remains effectively unchanged in practice.

Importantly, the empirical overhead remains below $0.003$ seconds on average, confirming that CAWI preserves the computational efficiency characteristic of randomized neural networks while improving stability and performance.

\end{document}